\pdfoutput=1

\documentclass[11pt]{article}

\usepackage[final]{acl}
\usepackage{amsmath}
\usepackage{times}
\usepackage{latexsym}
\usepackage{placeins}
\usepackage[T1]{fontenc}

\usepackage[utf8]{inputenc}

\usepackage{microtype}

\usepackage{inconsolata}
\usepackage{float}
\usepackage{graphicx}
\usepackage{amsfonts}
\usepackage{xcolor}
\usepackage{listings}
\usepackage{algorithm,algpseudocode}
\usepackage[most]{tcolorbox}
\usepackage{rotating}
\usepackage{booktabs}
\usepackage{subcaption}   
\usepackage{multirow}  
\usepackage{geometry}
\usepackage{tabularx}  
\usepackage{enumitem}
\usepackage{stfloats}
\usepackage{cuted}
\usepackage{colortbl}               
\usepackage{pgf}                    

\usepackage[colorinlistoftodos,prependcaption,textsize=tiny]{todonotes}
\title{ProCut: LLM Prompt Compression via Attribution Estimation}

\author{Zhentao Xu, Fengyi Li, Albert Chen, Xiaofeng Wang \\
LinkedIn Corporation \\ 
\texttt{\{zhexu, fenli, abchen, xiaofwang\}@linkedin.com}}

\begin{document}

\maketitle

\begin{abstract}

In large-scale industrial LLM systems, prompt templates often expand to thousands of tokens as teams iteratively incorporate sections such as task instructions, few-shot examples, and heuristic rules to enhance robustness and coverage. This expansion leads to bloated prompts that are difficult to maintain and incur significant inference latency and serving costs. To address this, we introduce \textit{\textbf{Pro}mpt \textbf{C}ompression via Attrib\textbf{u}tion Es\textbf{t}imation} (\textbf{ProCut}), a flexible, LLM-agnostic, training-free framework that compresses prompts through attribution analysis. ProCut segments prompt templates into semantically meaningful units, quantifies their impact on task performance, and prunes low-utility components. Through extensive experiments on five public benchmark datasets and real-world industrial prompts, we show that ProCut achieves substantial prompt size reductions (78\% fewer tokens in production) while maintaining or even slightly improving task performance (up to 62\% better than alternative methods). We further introduce an LLM-driven attribution estimator that reduces compression latency by over 50\%, and demonstrate that ProCut integrates seamlessly with existing prompt-optimization frameworks to produce concise, high-performing prompts.
\end{abstract}

\renewcommand{\thefootnote}{\fnsymbol{footnote}}
\footnotetext[1]{Preprint available at \href{https://arxiv.org/abs/2508.02053}{arXiv:2508.02053}.}
\renewcommand{\thefootnote}{\arabic{footnote}}

\section{Introduction}

Recent advances in Generative AI have brought large language models (LLMs) such as GPT-4 \citep{achiam2023gpt} and Claude \citep{anthropichome} into production pipelines for question answering, code generation, and retrieval-augmented search \citep{kamalloo2023evaluating, zan2023large, zhu2023large}. In industry, these systems are typically driven by prompt templates that grow organically over time as teams iteratively incorporate sections such as task instruction and heuristic rules. Few-shot in-context learning and chain-of-thought prompting magnify this expansion \citep{wei2022chain,brown2020language}, so prompts spanning several thousand tokens are now common \citep{hsieh2024automatic}.

Prompt bloat presents three primary challenges. The first is inflated inference latency and escalating API expenditure \citep{jiang2023llmlingua}; the second is degraded task accuracy, as vital instructions can be diluted or forgotten in very long prompts \citep{liu2024lost}. The third is mounting maintenance debt, with overlapping or conflicting sections making prompts increasingly hard to audit and debug. As a result, prompt compression has emerged as a crucial research and engineering focus. Prior work spans hard methods, such as Selective Context, LLMLingua, LLMLingua-2, Nano-Capsulator, Selection-p, DisComp, and EFPC \citep{li2023compressing, jiang2023llmlingua, pan2024llmlingua, chuang2024learning, chung2024selection, quancai2025discomp, cao2025efpc}, which perform token-level removal or generative paraphrasing, and soft methods, such as Gisting, AutoCompressor, ICAE, and 500×Compressor \citep{mu2023learning, chevalier2023adapting, ge2023context, li2024500xcompressor}, which encode prompts into compressed continuous embeddings. However, token-level methods can produce grammatically incorrect or disfluent text, complicating prompt maintenance and manual post-editing, whereas soft-embedding techniques lack cross-model generalizability and must be retrained for each new LLM, limiting their scalability across diverse production pipelines.

In this study, we introduce ProCut, a flexible, training-free framework that compresses prompt templates by leveraging attribution-estimation methods. Unlike token-level compression approaches, ProCut treats a prompt as a set of semantically coherent text segments, which are typically contiguous sentences or paragraphs, and casts prompt compression as a feature-selection problem, with each segment encoded as a binary feature. This formulation allows either direct use of established attribution techniques such as Shapley values (SHAP) \citep{lundberg2017unified}, Leave-One-Out (LOO) \citep{lei2018distribution}, LASSO regression \citep{tibshirani96regression}, or our newly proposed LLM-driven attribution method, to identify and retain the most impactful segments. We also demonstrate that ProCut integrates seamlessly with prompt-optimization frameworks such as TextGrad \citep{yuksekgonul2025optimizing}; by alternating between optimization and compression, the resulting prompts are both markedly more concise and more effective. In this paper, we make the following contributions:
\begin{enumerate}
\item We present the first use of attribution estimation for prompt compression, supporting diverse attribution algorithms.
\item We propose a constant-call, LLM-driven attribution estimator that preserves performance while significantly reducing latency.
\item We integrate prompt compression with prompt optimization frameworks, delivering concise and high-performing prompts.
\end{enumerate}

The software architecture and implementation details are in Appendix~\ref{ref_appendix_section_system_diagram} and~\ref{ref_section_user_interface}.

\section{Related Work} 

Automatic prompt optimization refines prompts to maximize LLM performance without direct model access \citep{ramnath2025systematic}. Prior work includes prompt rewriting \citep{li2024learning}, in-context example selection \citep{gupta2023coverage}, and gradient-based updates \citep{pryzant2023automatic, yuksekgonul2025optimizing}. Automatic prompt optimization often causes significant prompt growth; TextGrad, for instance, adds roughly 500 tokens per iteration (Section~\ref{ref_subsection_procut_as_regularization_layer}), raising execution costs and maintenance burdens \citep{das2025greater}.

Prompt compression aims to shorten prompts to improve inference efficiency and reduce costs. Existing approaches fall into two categories. Hard methods such as Selective Context, LLMLingua, LLMLingua-2, Nano-Capsulator, Selection-p, DisComp, and EFPC \citep{li2023compressing, jiang2023llmlingua, pan2024llmlingua, chuang2024learning, chung2024selection, quancai2025discomp, cao2025efpc} remove or paraphrase tokens but can break placeholders, produce disfluent text, and provide limited control over the final length. Soft methods such as Gisting, AutoCompressor, ICAE, and 500×Compressor \citep{mu2023learning, chevalier2023adapting, ge2023context, li2024500xcompressor} compress prompts into continuous embeddings; these representations lack interpretability and cross-model transferability and require retraining for each LLM.

\section{ProCut: Prompt Compression via Attribution Estimation}

\begin{figure*}
    \centering
    \includegraphics[width=0.99\linewidth]{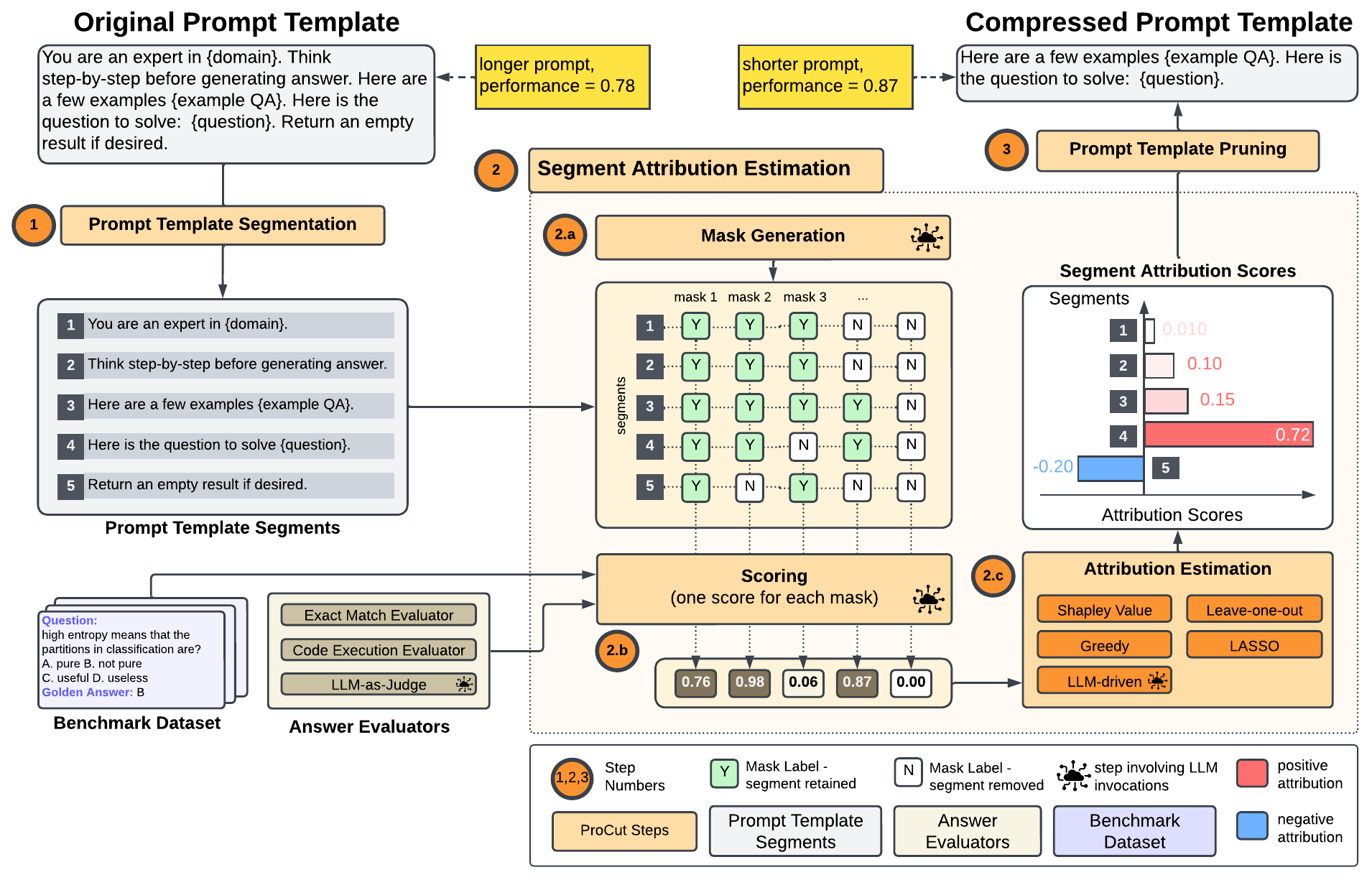}
    \caption{ProCut framework overview. The process consists of three stages: segmenting the prompt template, estimating the importance of each segment via attribution analysis, and pruning low-impact segments.}
    \label{ref_figure_procut_overall_method_diagram}
\end{figure*}

\subsection{Problem Formulation}
\label{ref_subsection_problem_formulation}
We assume an initial prompt template \(p\) whose placeholders are filled at inference time and that can be partitioned into an ordered list of \(M\) disjoint segments \([p_1,\dots,p_M]\). A ground-truth dataset \(\mathcal D=\{(x_i,y_i)\}\) provides inputs and reference outputs. For each input \(x_i\), the instantiated prompt \(p(x_i)\) is sent to a black-box LLM, yielding \(\hat y_i=\text{LLM}(p(x_i))\), which is evaluated with a task metric \(s(y_i,\hat y_i)\) or a reference-free metric \(s(\hat y_i)\). The goal is to choose a subset \(\mathcal K\subseteq\{1,\dots,M\}\) of size \(k\) so that the compressed template \(p_{\mathcal K}\) maximizes the average metric value on \(\mathcal D\).

ProCut compresses prompts in three steps: \textbf{prompt template segmentation} (Sec.~\ref{ref_subsection_prompt_template_segmentation}) divides the template into segments; \textbf{segment attribution estimation} (Sec.~\ref{ref_subsection_prompt_segment_attribution_estimation}) scores each segment; and \textbf{prompt template pruning} (Sec.~\ref{ref_subsection_prompt_pruning}) removes low-impact segments to produce a compact prompt template.

\subsection{Prompt Template Segmentation}
\label{ref_subsection_prompt_template_segmentation}

The prompt template segmentation module decomposes a given prompt template \(p\) into \(M\) segments \([p_1,\dots,p_M]\) where $M$ denotes the number of segments and is a configurable hyperparameter. We employ three  strategies: \textbf{(a) pre-defined segmentation}, in which domain owners label logical blocks for targeted compression; \textbf{(b) structure-aware segmentation}, which cuts at natural sentence or paragraph boundaries, assuming these units correspond to semantically coherent components; and \textbf{(c) LLM-driven segmentation}, which prompts LLM to partition unstructured or model-generated templates such as those produced by TextGrad. Figure~\ref{ref_prompt_for_prompt_segmentation} illustrates the prompt used in the LLM-driven segmentation method, and Appendix~\ref{ref_appendix_example_segmentation} provides examples of three segmentation methods.

\subsection{Segment Attribution Estimation} 
\label{ref_subsection_prompt_segment_attribution_estimation}

\paragraph{Perturbation-based Attribution Methods} The attribution module accepts a segmented prompt template \([p_1,\dots,p_M]\) and produces scores \([a_1,\dots,a_M]\) that quantify each segment’s contribution to the task metric. We rely on perturbation-based methods, which require only API access and thus keep the framework model-agnostic. Any perturbation scheme can be plugged in; for illustration we report four representative algorithms: \textbf{(a) Shapley values (SHAP)}, which estimates marginal contributions via Monte Carlo subsets; \textbf{(b) Leave-One-Out (LOO)}, which measures the performance drop when a single segment is removed; \textbf{(c) LASSO regression}, which fits a sparse linear model to scores from randomly masked prompts; and \textbf{(d) greedy forward selection}, which adds segments sequentially based on observed gain. All methods are evaluated on a held-out test set \(\mathcal{D}_{\text{test}}\), using the task-specific metric \(s(y, \hat{y})\) to assess prediction quality.

\paragraph{LLM-driven Attribution Estimation} \label{ref_subsection_llm_driven_attribution_analysis} The attribution methods discussed above require a large number of LLM invocations, ranging from \(M\) for Leave-One-Out to \(2^{M}\) for Shapley values, making them costly and impractical for large-scale deployment, especially when the number of segments and dataset size are high. To address this limitation, we propose an \emph{LLM-driven} approach (Algorithm~\ref{ref_algorithm_llm_driven_attribution_estimation}) that leverages the model’s own semantic understanding of the prompt. Specifically, rather than relying on a single zero-shot estimate \citep{jeong2024llm}, our method prompts the LLM to generate a bounded set of candidate masks that highlight segments it deems important. Each mask is evaluated on the training set, and the resulting feedback is returned to the LLM, forming a probe-and-test loop that refines the segment rankings iteratively. By limiting the number of candidate masks, the process completes in fewer than \(M\) LLM calls, substantially cheaper than traditional black-box techniques while maintaining high attribution fidelity.

\subsection{Prompt Template Pruning}
\label{ref_subsection_prompt_pruning}
After scoring the segments, we prune the prompt template by retaining the top \(\lfloor rM \rfloor\) segments, where \(M\) is the total number of segments and \(r \in [0,1]\) is a user-defined compression ratio. Retained segments preserve their original order to maintain context. The ratio \(r\) can be fixed to meet latency or cost targets or tuned for task performance.

\section{Experiments}
We conduct a series of experiments to evaluate the effectiveness and efficiency of ProCut across diverse settings. Specifically, we aim to answer the following research questions (RQs):
\begin{itemize}
\item \textbf{RQ1}: How effective is ProCut in compressing prompts compared to baseline methods?
\item \textbf{RQ2}:  Can LLMs' semantic understanding capabilities be leveraged to accelerate prompt attribution analysis while maintaining high attribution quality?
\item \textbf{RQ3}: How can ProCut be integrated with automated prompt optimization frameworks such as TextGrad for generating high-performing and efficient prompts?
\end{itemize}

\subsection{Datasets}
\label{ref_subsection_dataset}

\begin{table}[t]
\centering
\scriptsize
\setlength{\tabcolsep}{3pt}
\begin{tabular}{p{1.2cm} p{2.1cm} c c p{1.4cm}}
\toprule
\textbf{Dataset} & \textbf{Tasks} & \textbf{Train} & \textbf{Test} & \textbf{Metric} \\
\midrule
GSM8K     & -- & 20 & 100 & Exact Match \\
SQuAD     & -- & 20 & 100 & F1-score \\
HumanEval & -- & 20 & 100 & pass@1 \\
\midrule
\multirow{4}{*}{BBH} 
& Geometry Shapes     & 20 & 100 & \multirow{4}{*}{Exact Match} \\
& Object Counting     & 20 & 100 &  \\
& Color Reasoning     & 20 & 100 &  \\
& Penguins            & 20 & 100 &  \\
& Temporal Sequence   & 20 & 100 &  \\
\midrule
\multirow{4}{*}{MMLU} 
& College Medicine    & 20 & 100 & \multirow{4}{*}{Exact Match} \\
& College Math        & 20 & 75  &  \\
& Anatomy             & 20 & 100 &  \\
& Astronomy           & 20 & 100 &  \\
\bottomrule
\end{tabular}
\caption{Summary of datasets, tasks, splits, and metrics.}
\label{ref_table_dataset_and_metric_summary}
\end{table}

To comprehensively assess ProCut’s compression effectiveness and attribution accuracy, we evaluate it on 12 tasks spanning five benchmark datasets across four diverse categories (Table~\ref{ref_table_dataset_and_metric_summary}): mathematical reasoning with \textbf{GSM8K}~\cite{cobbe2021training}; code generation with \textbf{HumanEval}~\cite{chen2021evaluating}; extractive QA with \textbf{SQuAD}~\cite{rajpurkar2016squad}; and broad knowledge and reasoning from nine tasks sampled from \textbf{BBH}~\cite{suzgun2023challenging} and \textbf{MMLU}~\cite{hendrycks2020measuring}. To ensure a fair and reliable assessment, we use each dataset’s official train/test split if available; when subsampling is required, we uniformly sample within each split.

\subsection{Evaluation Metrics}
\label{ref_subsection_evaluation}
Table~\ref{ref_table_dataset_and_metric_summary} lists the metrics used. Following prior work, we report Exact Match for GSM8K, BBH, and MMLU \citep{cobbe2021training,suzgun2023challenging,fu2023estimating}; unbiased Pass@1 for HumanEval \citep{chen2021evaluating}; and F1 score for SQuAD \citep{li2024500xcompressor}.

\subsection{Prompt Template}
\label{subsection_prompt_template}

To ensure that our experiments start with representative and well-engineered prompts, we constructed an initial template with five widely adopted segment types: role-playing, zero-shot chain-of-thought prompting, few-shot chain-of-thought examples, a question placeholder, and a context placeholder. These segments reflect best practices in prompt design: few-shot examples are taken directly from the BBH, GSM8K, and MMLU datasets; the zero-shot cue follows \citet{kojima2022large}; and role-playing uses the common pattern \texttt{``You are an expert in \{domain\}''}. Segments are instantiated only when relevant (the context placeholder appears only in SQuAD, and few-shot examples only in GSM8K, BBH, and MMLU). The complete template used in the RQ1 and RQ2 experiments is provided in Table~\ref{ref_table_prompt_template} of Appendix~\ref{ref_appendix_compression_initial_prompt_template}.

\subsection{Implementation Details}

We use the April 2025 release of GPT-4.1 mini model via the OpenAI API.\footnote{\url{https://platform.openai.com/}} To improve the stability of outputs produced by the LLM, we set the temperature to 0 and allow a 4000-token context window for long TextGrad prompts. Experiments run on a MacBook Pro (M3 Pro, 36 GB RAM) with Python 3.12, issuing API calls in parallel through ten \texttt{concurrent.futures} threads. All algorithms keep their default hyperparameters. For RQ1 and RQ2 we apply the template from Section~\ref{subsection_prompt_template} at compression ratios \(r\in\{25\%,50\%,75\%\}\); for RQ3 we use LLM-driven segmentation (up to five segments) with \(r\in\{40\%,60\%,80\%\}\). To ensure reliable evaluation, each setting is run for five iterations, and the averaged results are reported.

\subsection{Prompt Compression Performance (RQ1)}

We compare ProCut with four segment-level prompt compression baselines: (i) \emph{Vanilla LLM}, where the LLM is instructed to compress the prompt \citep{pu2024style}; (ii) \emph{Selective Context} \citep{li2023compressing}, a heuristic filter that removes low-utility segments; (iii) \emph{Random Selection} \citep{jiang2023llmlingua}; and (iv) a \emph{Brute-force Oracle} that exhaustively searches all \(2^{M}\) segment subsets. 
We further compare ProCut with competitive token-level compression methods, including (v) \emph{LLMLingua} and \emph{LLMLingua-2} \citep{jiang2023llmlingua,pan2024llmlingua}; we address the challenge of prompt template placeholder corruption by manually repairing them while keeping all other compressed tokens unchanged.

\begin{figure}[ht] 
  \centering
  \includegraphics[width=\columnwidth]{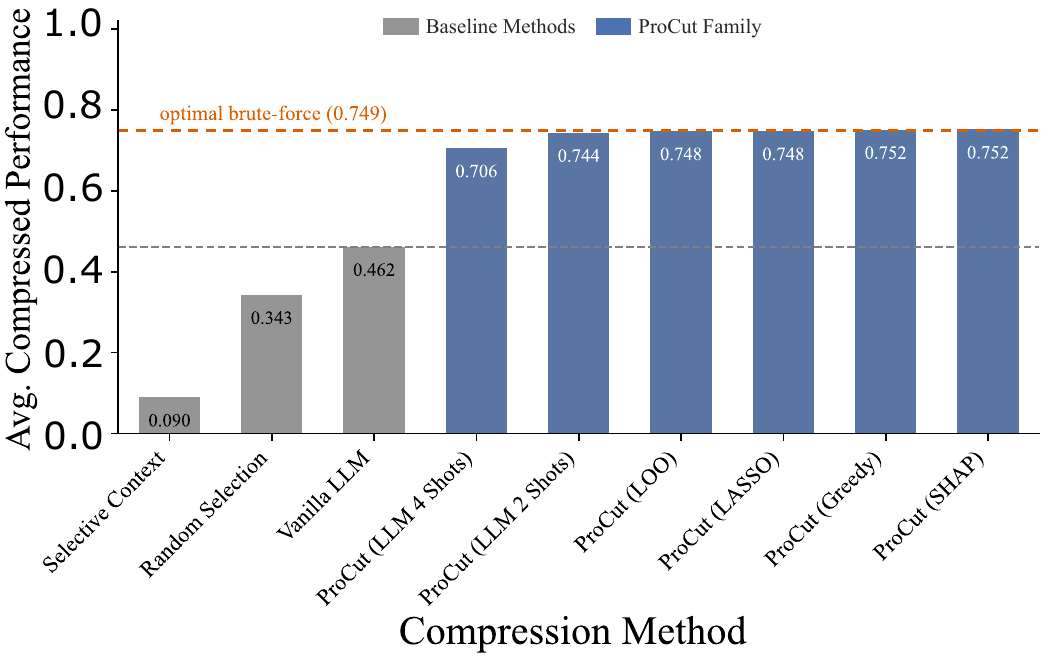}
\caption{Performance of compressed prompts. Grey bars show baselines, blue bars show ProCut variants, the dashed orange line indicates the brute-force oracle.}

  \label{ref_fig_average_compression_performance}
\end{figure}

Figure~\ref{ref_fig_average_compression_performance} and Table~\ref{ref_table_average_compression_performance_compression_ratio_level} summarize the performance of compressed prompts for ProCut and the segment-level compression baselines. 
ProCut consistently outperforms all baselines and closely approaches the brute-force oracle that exhaustively searches all \(2^{M}\) segment subsets. 
On average, it improves compressed prompt performance from 0.46 to over 0.70, confirming that attribution-guided pruning effectively retains segments most critical for task quality. Table~\ref{tab:token_level_method_comparison} further reports the comparison with competitive token-level methods \emph{LLMLingua} and \emph{LLMLingua-2} on the SQuAD dataset. ProCut achieves substantially higher performance at moderate and low compression levels, and remains competitive under high compression, indicating that segment-level attribution effectively preserves essential content even with strong compression.

\begin{table}[ht!]
\centering
\scriptsize                                    
\setlength{\tabcolsep}{4pt}                    
\begin{tabular}{lcccc}
\specialrule{1.2pt}{1pt}{1pt}

  & \multicolumn{3}{c}{\textbf{Compression Ratio}} 
  &  \\
\cmidrule(lr){2-4}
\textbf{Method} & \textbf{25\%} & \textbf{50\%} & \textbf{75\%} & \textbf{Average} \\      
\specialrule{1.2pt}{1pt}{1pt}
Random Selection (Baseline) & 0.104 & 0.359 & 0.567 & 0.343 \\
Vanilla LLM (Baseline) & 0.064 & 0.579 & 0.743 & 0.462 \\
Selective Context (Baseline) & 0.041 & 0.122 & 0.106 & 0.090 \\
Brute Force (Oracle) & 0.567 & 0.847 & 0.834 & 0.749 \\
\midrule
ProCut (SHAP) & \textbf{0.575} & 0.841 & \textbf{0.841} & \textbf{0.752} \\
ProCut (Leave-One-Out) & 0.570 & 0.839 & 0.836 & 0.748 \\
ProCut (LASSO) & 0.573 & 0.846 & 0.826 & 0.748 \\
ProCut (Greedy Forward) & 0.569 & \textbf{0.848} & 0.838 & \textbf{0.752} \\
ProCut (LLM-as-Ranker 2 Shots) & 0.570 & 0.837 & 0.826 & 0.744 \\
ProCut (LLM-as-Ranker 4 Shots) & 0.561 & 0.776 & 0.781 & 0.706 \\

\specialrule{1.2pt}{1pt}{1pt}
\end{tabular}
\caption{Average performance of compressed prompts across compression ratios. The uncompressed prompt template achieves a performance of 0.817. Detailed breakdowns are provided in Appendix~\ref{ref_appendix_compression_performance_task_level},~\ref{ref_appendix_compression_performance_dataset_level}, and~\ref{ref_appendix_compression_performance_aggregated}.}
\label{ref_table_average_compression_performance_compression_ratio_level}
\end{table}

\begin{table}[ht!]
\centering
\scriptsize                                    
\setlength{\tabcolsep}{4pt}                    
\begin{tabular}{lcccc}
\specialrule{1.2pt}{1pt}{1pt}
 & \multicolumn{3}{c}{\textbf{Compression Ratio}} & \\
\cmidrule(lr){2-4}
\textbf{Method} & \textbf{25\%} & \textbf{50\%} & \textbf{75\%} & \textbf{Average} \\
\hline
LLMLingua (Baseline) & 0.228 & 0.236 & 0.201 & 0.222 \\
LLMLingua-2 (Baseline) & 0.019 & 0.058 & \textbf{0.755} & 0.278 \\
\midrule
ProCut (average across all variants) & \textbf{0.240} & \textbf{0.832} & 0.736 & \textbf{0.603} \\
\specialrule{1.2pt}{1pt}{1pt}
\end{tabular}
\caption{Comparison of ProCut and token-level compression methods on the SQuAD dataset. The corresponding compressed prompt templates are provided in Appendix~\ref{ref_appendix_compressed_prompt_template}.}
\label{tab:token_level_method_comparison}
\end{table}

\begin{figure}[ht]
    \centering
    \includegraphics[width=0.99\linewidth]{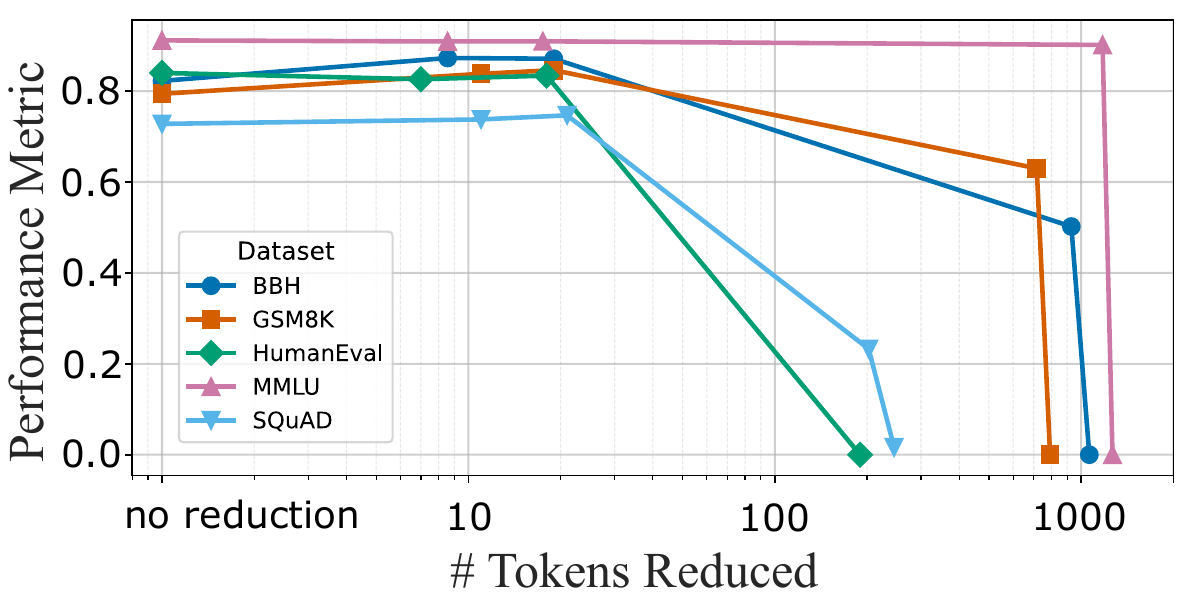}
    \caption{Prompt Performance vs Token Reductions. }
    \label{ref_fig_3_prompt_performance_vs_token_reduction}
\end{figure}

Interpretability plays a critical role in industrial prompt compression and optimization. ProCut natively supports this through segment-level attribution analysis (Table~\ref{ref_table_exp_2_attribution_results_dataset_level}) and performance–token reduction trade-off visualization (Figure~\ref{ref_fig_3_prompt_performance_vs_token_reduction}). For example, the attribution results over the open-source datasets (Table~\ref{ref_table_dataset_and_metric_summary}) show that \emph{few-shot CoT examples} contribute significantly to performance, underscoring the value of curated demonstrations, while \emph{zero-shot CoT} and \emph{role-playing} have minimal impact, suggesting that modern LLMs may already internalize such heuristics. The trade-off plot in Figure~\ref{ref_fig_3_prompt_performance_vs_token_reduction} provides actionable insight into how token reduction impacts performance, helping practitioners strike an optimal balance between efficiency and quality. Importantly, picking the ``sweet point'' in this trade-off plot is use-case-specific: it is shaped by both the product’s performance requirements and the available cost resources.

\begin{table}[ht!]
\centering
\scriptsize
\setlength{\tabcolsep}{2.8pt}
\begin{tabular}{lccccc}
\toprule
\textbf{Method} & \textbf{GSM8K} & \textbf{SQuAD} & \textbf{HumanEval} & \textbf{BBH} & \textbf{MMLU} \\
\midrule
Role-playing & 
  \cellcolor[HTML]{FFFEFE}\makebox[0pt][c]{0.009} & 
  \cellcolor[HTML]{FFFEFE}\makebox[0pt][c]{0.011} & 
  \cellcolor[HTML]{FFFFFF}\makebox[0pt][c]{0.000} & 
  \cellcolor[HTML]{EFF4FA}\makebox[0pt][c]{-0.019} & 
  \cellcolor[HTML]{E5EDF6}\makebox[0pt][c]{-0.031} \\
Zero-shot CoT & 
  \cellcolor[HTML]{F0F4FA}\makebox[0pt][c]{-0.018} & 
  \cellcolor[HTML]{FDFEFE}\makebox[0pt][c]{-0.002} & 
  \cellcolor[HTML]{FFFFFF}\makebox[0pt][c]{0.000} & 
  \cellcolor[HTML]{FFFDFD}\makebox[0pt][c]{0.014} & 
  \cellcolor[HTML]{E2EBF5}\makebox[0pt][c]{-0.035} \\
Few-shot CoT Examples & 
  \cellcolor[HTML]{FFE6E6}\makebox[0pt][c]{\textbf{0.199}} & 
  -- & 
  -- & 
  \cellcolor[HTML]{FFEBEB}\makebox[0pt][c]{\textbf{0.158}} & 
  \cellcolor[HTML]{FFF5F5}\makebox[0pt][c]{0.079} \\
Context Placeholder & 
  -- & 
  \cellcolor[HTML]{FFE1E1}\makebox[0pt][c]{\textbf{0.232}} & 
  -- & 
  -- & 
  -- \\
Question Placeholder & 
  \cellcolor[HTML]{FF9A9A}\makebox[0pt][c]{\textbf{0.789}} & 
  \cellcolor[HTML]{FFBDBD}\makebox[0pt][c]{\textbf{0.519}} & 
  \cellcolor[HTML]{FF8080}\makebox[0pt][c]{\textbf{1.000}} & 
  \cellcolor[HTML]{FFAEAE}\makebox[0pt][c]{\textbf{0.637}} & 
  \cellcolor[HTML]{FF9C9C}\makebox[0pt][c]{\textbf{0.777}} \\
\bottomrule
\end{tabular}
\caption{Attribution of prompt components. Bold: $|\text{attribution}| > 0.1$; red/blue: positive/negative impact.}
\label{ref_table_exp_2_attribution_results_dataset_level}
\end{table}

\begin{figure*}[ht]
    \centering
    \includegraphics[width=1\linewidth]{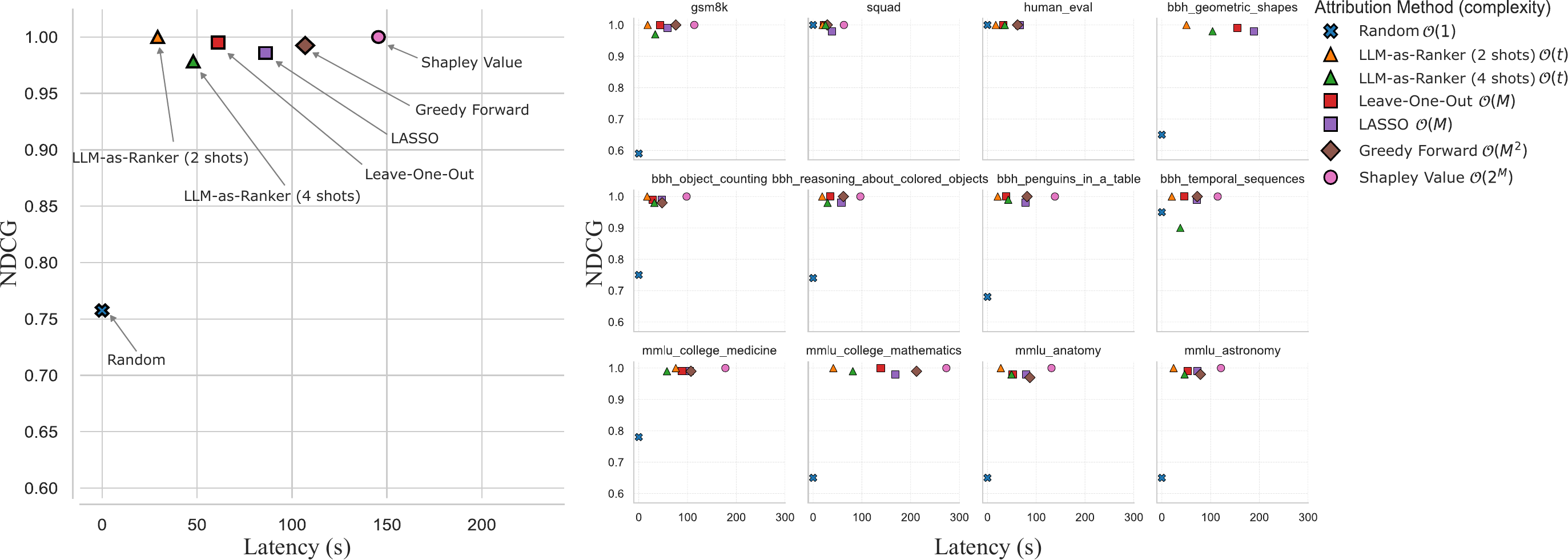}
\caption{Trade-off between attribution quality (NDCG) and computational cost (latency in seconds). The left plot shows the average quality vs. cost across all datasets, while the right plot presents results for each individual task.}
    \label{fig:attr_latency_tradeoff}
\end{figure*}

\begin{figure}[ht!]
    \centering
    \includegraphics[width=0.99\linewidth]{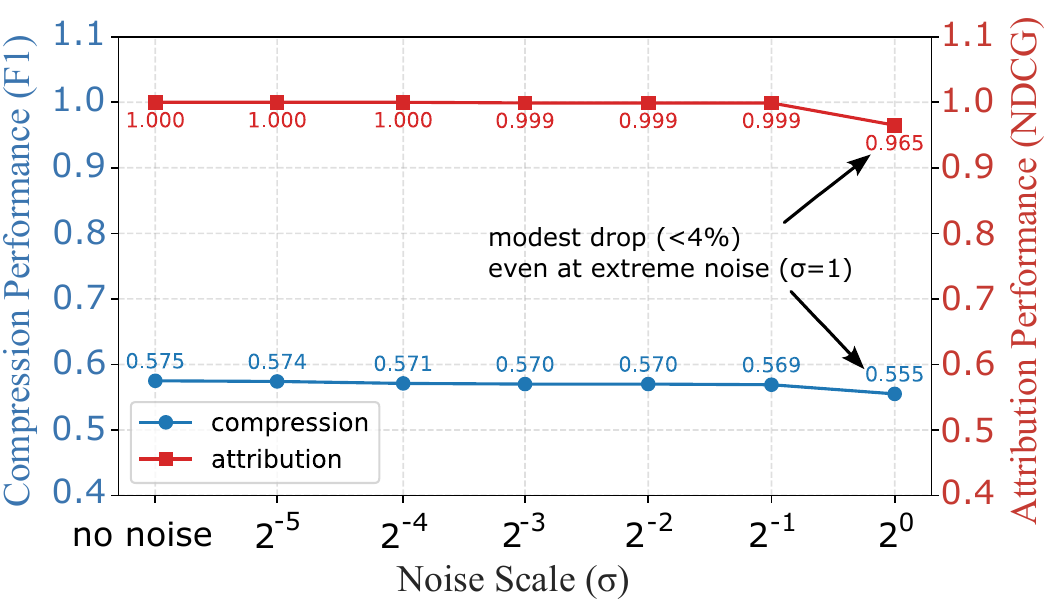}
    \caption{Robustness of ProCut on the SQuAD dataset with noisy metrics: compression performance (F1 of compressed prompts) and attribution performance (NDCG vs.\ noise-free reference) remain stable, with only modest degradation under large-scale noise.}  
    \label{ref_fig_4_prompt_performance_vs_noiseness}
\end{figure}

We evaluated ProCut's robustness under noisy and weak metrics, a common scenario in production where labels are costly and golden metrics are hard to define early on. In the synthetic noise experiment on SQuAD (Figure~\ref{ref_fig_4_prompt_performance_vs_noiseness}), Gaussian noise was injected into the evaluation metric; ProCut remained stable under moderate noise ($\leq1\%$ variation for $\sigma \leq 0.5$) and showed only a modest drop at extreme noise. In the weak supervision experiment on GSM8K, replacing gold labels with label-free LLM-as-judge scores yielded nearly identical performance ($\sim$0.90 accuracy) and near-perfect attribution alignment ($\text{NDCG} \approx 1$). These results demonstrate ProCut's robustness to noisy and weak metrics.

\subsection{LLM-driven Attribution Estimation (RQ2)}

We benchmark our LLM-driven attribution estimator against four commonly used black-box methods introduced in Section~\ref{ref_subsection_prompt_segment_attribution_estimation}. Figure~\ref{fig:attr_latency_tradeoff} compares these methods along two axes: attribution fidelity in NDCG and runtime in seconds. Our LLM-driven estimator achieves comparable performance to the SHAP “gold standard” while substantially reducing runtime. The LLM-as-Ranker (2-shot) variant lowers end-to-end latency by 80\%, 52\%, and 66\% compared to SHAP, LOO, and LASSO respectively, while maintaining near-identical fidelity. This shifts the latency–fidelity trade-off curve significantly, showing that LLMs can effectively estimate attribution with minimal computational overhead by leveraging their semantic reasoning and ranking capabilities.

\subsection{Integrating ProCut into Prompt Optimization (RQ3)}
\label{ref_subsection_procut_as_regularization_layer}

\begin{figure}[ht]
    \centering
    \includegraphics[width=0.99\linewidth]{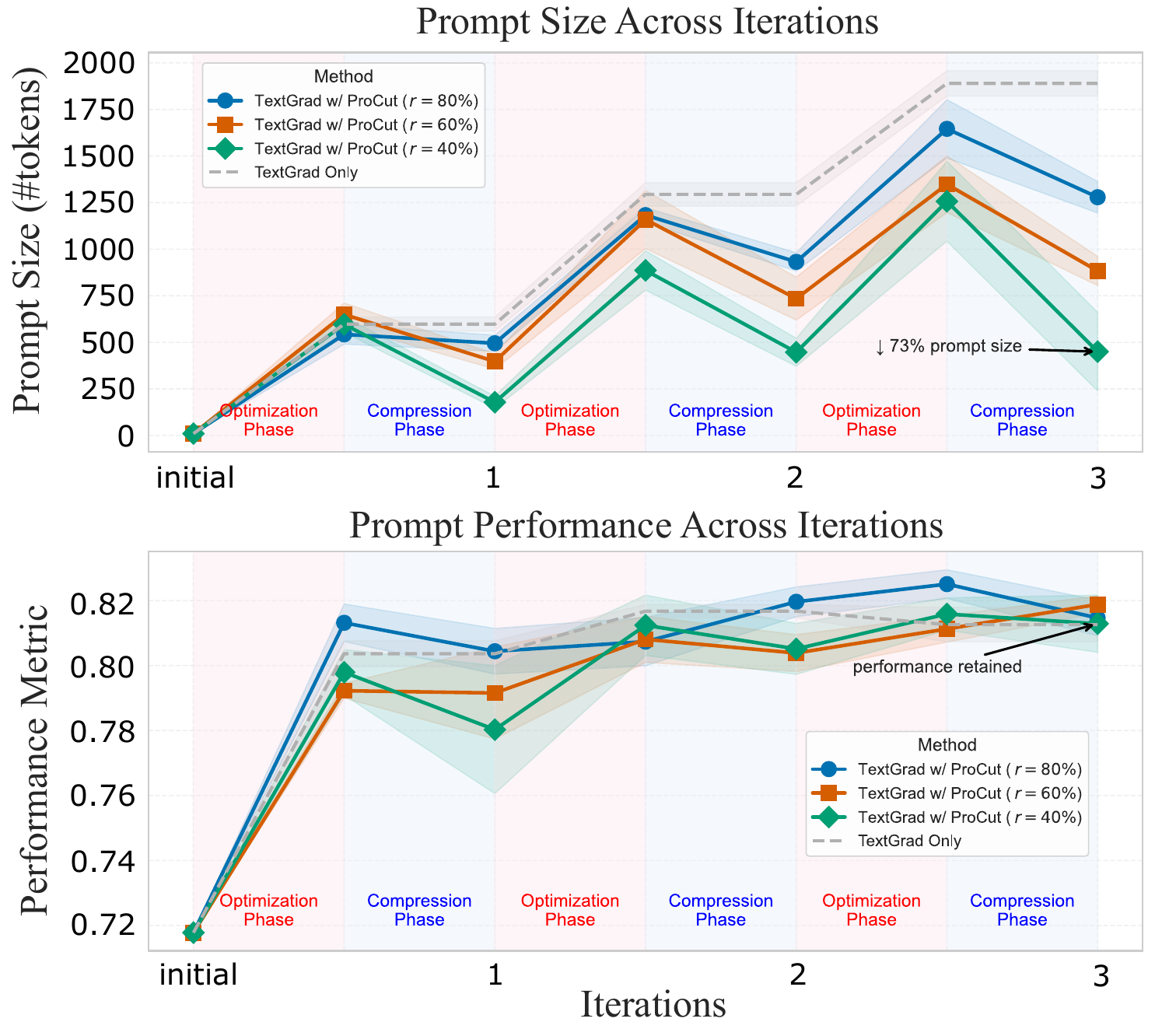}
    \caption{Prompt template size and performance across TextGrad + ProCut iterations on the SQuAD dataset. We use the prompt template \texttt{``Please finish the extractive question answering task''} as the initial prompt, and report the size and performance after each iteration under compression ratios $\in \{0.4, 0.6, 0.8\}$.}
    \label{ref_fig_exp_4_textgrad_procut}
\end{figure}

Figure~\ref{ref_fig_exp_4_textgrad_procut} compares pure TextGrad optimization with a ProCut-regularized variant on SQuAD dataset, where prompt compression is applied after each iteration. We observe that integrating ProCut into the optimization loop effectively controls prompt length growth without sacrificing performance. Specifically, after three iterations, the ProCut-regularized groups generate final prompts with only 27\%, 47\%, and 66\% of the token count compared to the TextGrad-only group, under compression ratios of 40\%, 60\%, and 80\%, respectively—while maintaining comparable performance of 0.815, 0.819, and 0.813, close to the uncompressed baseline of 0.813. See Appendix \ref{ref_appendix_textgrad_and_procut_prompt_examples} for exemplary prompts from both groups.

\section {Production Use Cases}

We evaluated ProCut using real-world prompts from two high-traffic production pipelines: intent classification and candidate qualification assessment. These use cases demonstrate ProCut’s ability to significantly reduce prompt length without compromising accuracy, resulting in substantial LLM inference cost savings at scale (Table~\ref{ref_table_production_usecase}).

\begin{table}[ht]
\centering
\scriptsize
\begin{tabular}{
    >{\centering\arraybackslash}p{1.4cm} 
    >{\centering\arraybackslash}p{1.4cm} 
    >{\centering\arraybackslash}p{1.4cm} 
    >{\centering\arraybackslash}p{1.75cm}
}
\toprule
\textbf{Use Case} & \textbf{Token Reduction} & \textbf{Cost Saved per 1M Calls} & \textbf{Performance Impact} \\
\midrule
Intent Cls.    & 73\% & \textasciitilde\$7K & Preserved \\
Qual. Assess.  & 84\% & \textasciitilde\$8K & Slightly improved \\
\bottomrule
\end{tabular}
\caption[ProCut performance on production prompts]{ProCut performance on production prompts.\footnotemark}
\label{ref_table_production_usecase}
\end{table}
\footnotetext{Costs are estimated based on latest publicly available \href{https://openai.com/api/pricing}{GPT-4o API pricing}; actual deployment costs may vary.}

\paragraph{ProCut for Intent Classification Prompt}
ProCut was applied to a production prompt used to classify recruiter actions into predefined intent categories based on a combination of structured metadata and free-text signals. Using a human-annotated dataset and classification accuracy as the evaluation metric, ProCut achieved a 73\% reduction in prompt length with no loss in accuracy.

\paragraph{ProCut for Qualification Assessment Prompt}  
ProCut was also evaluated on a qualification assessment task with 46 human-labeled examples. The original prompt, exceeding 2{,}200 tokens, was reduced to approximately 300 tokens (84\% reduction). Performance was slightly improved post-compression, indicating that ProCut preserves essential information while eliminating redundant context.

\section {Conclusions}

In this paper, we introduced ProCut, a flexible, LLM-agnostic, training-free prompt compression framework that formulates template pruning as segment-level attribution. Benchmarked across 12 tasks from five representative datasets, ProCut achieves an average 62\% performance gain over strong compression baselines (Figure~\ref{ref_fig_average_compression_performance}). We extend classical attribution methods by incorporating an LLM-driven variant that reduces the computational cost from $\Omega(M)$ to a constant number of LLM calls, resulting in a 52\% reduction in runtime latency (Figure~\ref{fig:attr_latency_tradeoff}). We further show that ProCut integrates seamlessly with prompt optimization frameworks and produces prompts with only 27\% of the token count while achieving similar task performance (Figure~\ref{ref_fig_exp_4_textgrad_procut}). ProCut has also been applied to two production prompts, achieving 73\% and 84\% token reductions, respectively, leading to substantial cost savings in LLM inference (Table~\ref{ref_table_production_usecase}).

\section {Limitations}
While ProCut demonstrates strong empirical performance and practical applicability, several aspects warrant further investigation. First, ProCut assumes the availability of a reliable and directional evaluation metric to guide segment attribution. Although we have shown that ProCut remains robust under noisy and weak supervision signals, broader evaluation across diverse tasks would further validate its generality. Second, while our LLM-driven attribution estimator significantly reduces model invocation costs, it remains a heuristic that relies on the model’s introspective capabilities, which may be less reliable when prompts contain ambiguous or adversarial segments. Finally, our evaluation spans five benchmark datasets, some of which may overlap with the pretraining data of foundation models. Although our internal case studies provide preliminary evidence of generalization to real-world, out-of-distribution prompts, future work should systematically evaluate ProCut across more diverse and truly unseen domains.

\bibliography{main}
\clearpage
\appendix

\section{ProCut Prompts}
\subsection{Prompt for Prompt Segmentation}
Instructions for Prompt Segmentation. The instruction for prompt compression is shown in Figure~\ref{ref_prompt_for_prompt_segmentation}. When using the instruction, one needs to add the initial prompt into the \texttt{\{current\_prompt\}} placeholder.

\begin{figure}[ht]
\begin{tcolorbox}[
    enhanced,
    breakable,
    sharp corners=south,
    title=Prompt for Prompt Segmentation,
    fonttitle=\bfseries,
    colback=white,
    colframe=black,
    left=1pt,
    right=1pt,
    top=1pt,
    bottom=1pt,
    boxsep=1pt
]

\begin{lstlisting}[
    breaklines=true,
    basicstyle=\ttfamily\scriptsize,
    aboveskip=2pt,
    belowskip=2pt
]

Below is the prompt you need to split:
<prompt_start_below (this is not part of prompt)>
{current_prompt}
<prompt_end_above (this is not part of prompt)>

Instructions:
1. Split the prompt into at most {max_units} units.
2. Each unit should ideally represent a complete sentence or paragraph that preserves the original meaning.

Ensure:
1. No overlap between units.
2. No missing information from the original prompt.
3. Units can be concatenated to reconstruct the original prompt exactly.
4. No modifications, including punctuation, to the original content.
5. Placeholders (curly braces) must remain intact and within a single unit.
6. Logically-related content should be grouped together (again, you don't need to split the prompt into maximum number of units {max_units} if it doesn't make sense).

Example:
   Original prompt: "Please answer the following question <question>{{question}}</question>."
   Result: {{"units": [{{"template": "Please answer the following question <question>{{question}}</question>."}}]}}

Example 2:
   Original prompt: "Please answer the following question <question>{{question}}</question>.   Please provide your answer in the following format: <format>{{format}}</format>."
   Result: {{"units": [{{"template": "Please answer the following question <question>{{question}}</question>."}}, {{"template": "Please provide your answer in the following format: <format>{{format}}</format>."}}]}}

Please return the result in the following format:
{{
   "units": [
       {{
           "template": "unit 1 template"
       }},
       {{
           "template": "unit 2 template"
       }},
       ...
   ]
}}
\end{lstlisting}
\end{tcolorbox}
\caption{Prompt for Prompt Segmentation}
\label{ref_prompt_for_prompt_segmentation}
\end{figure}

\newpage
\subsection{Prompt for Generating Masks for Attribution
Estimation}
Instructions for generating masks for attribution estimation. The instruction for prompt compression is shown in Figure~\ref{ref_prompt_for_generating_mask_for_attribution_analysis}. When using the instruction, one needs to instantiate the prompt by plugging values, in specific, adding the segmented prompts into placeholder \texttt{\{segmented\_prompt\_template\}}, and providing the expected number of masks into placeholder \texttt{\{num\_mask\}}.

\begin{figure}[ht]
\begin{tcolorbox}[
    enhanced,
    breakable,
    sharp corners=south,
    title=Prompt for Generating Mask for Attribution Estimation,
    fonttitle=\bfseries,
    colback=white,
    colframe=black,
    left=1pt,
    right=1pt,
    top=1pt,
    bottom=1pt,
    boxsep=1pt
]

\begin{lstlisting}[
    breaklines=true,
    basicstyle=\ttfamily\scriptsize,
    aboveskip=2pt,
    belowskip=2pt
]

Below is a prompt that has already been segmented into text unit: {segmented_prompt_template}
I would like to select some components and test it on a dataset and then estimate the importance of each component using LLM.
Please read the segmented prompt and based on the semantic meaning,
choose {num_mask} masks that can help me gather more information in terms of estimating the importance of each component.
Output Format Instruction:

Please return only the JSON object of the following format:
{{
    "masks": [List[List[int]]] 
    "rationale": str
}}
- Field "masks": selected masks that can help gather more information about estimating the importance of each component.
    Each mask must be of the same length as {num_features}.
- Field "rationale": a string explaining the rationale behind the selection.

Example:
masks = [[0,0,1,0,0,1], [1,1,1,0,0,1], [0,0,0,0,0,1], [1,0,1,0,1,1]] represents 4 masks, 
with 1 representing the prompt component is selected and 0 representing the prompt component is not selected.

If you need to output more, then please put the JSON object within "```json" and "```". Please ensure the json object is valid (e.g. no # comments within the json object).

\end{lstlisting}

\end{tcolorbox}
\caption{\texttt{AskLLMForIndex}: Prompt for Generating Mask for Attribution Estimation}
\label{ref_prompt_for_generating_mask_for_attribution_analysis}
\end{figure}

\newpage
\subsection{Prompt for Estimating Attribution}
Instructions for estimating segment attributions. The instruction for prompt compression is shown in Figure~\ref{ref_prompt_for_estimating_attribution_ranking}. When using the instruction, one needs to add the experiment results (i.e. masks and corresponding performance) into the placeholder \texttt{\{experiments\}}.

\begin{figure}[ht]
\begin{tcolorbox}[
    enhanced,
    breakable,
    sharp corners=south,
    title=Prompt for Estimating Attribution Ranking,
    fonttitle=\bfseries,
    colback=white,
    colframe=black,
    left=1pt,
    right=1pt,
    top=1pt,
    bottom=1pt,
    boxsep=1pt
]
\begin{lstlisting}[
    breaklines=true,
    basicstyle=\ttfamily\scriptsize,
    aboveskip=2pt,
    belowskip=2pt
]
Below are the results with different combinations of prompt components and the corresponding correctness.  Please use the this information to determine which prompt components are important. 
{experiments}

Output Format Instruction:

Please return only the JSON object of the following format:
{{
    "ranking": List[int]
    "rationale": str,
}}

Note:
- Field "ranking": a list of integers representing the ranking of each feature according to its importance. The more important component should be ranked in front.
- Field "rationale": a string explaining the rationale behind the ranking.

Example:
ranking: [3, 4, 0, 2, 5, 1] # the most important component is the third (#3) component, the second important is the fourth (#4) component, etc. Please start counting from 0.

If you need to output more, then please put the JSON object within "```json" and "```". Please ensure the JSON object is valid (e.g. no # comments within the JSON object).

\end{lstlisting}
\end{tcolorbox}
\caption{\texttt{RankPrompt}: Prompt for Estimating Attribution Ranking}
\label{ref_prompt_for_estimating_attribution_ranking}
\end{figure}

\newpage
\section{Prompt Template Segmentation Example}
\label{ref_appendix_example_segmentation}

\begin{figure}[ht]
\begin{tcolorbox}[
    enhanced,
    breakable,
    sharp corners=south,
    title={Example Prompt Template Before Segmentation},
    fonttitle=\bfseries,
    colback=white,
    colframe=black,
    left=1pt,
    right=1pt,
    top=1pt,
    bottom=1pt,
    boxsep=1pt
]
\begin{lstlisting}[
    breaklines=true,
    basicstyle=\ttfamily\scriptsize,
    aboveskip=2pt,
    belowskip=2pt
]
You are an expert in climate change and environmental policy. Please read the following passage carefully. Then, summarize the main arguments presented in the passage. After that, provide at least three potential counterarguments. Next, identify which of the arguments are supported by scientific evidence. Please also highlight any logical fallacies present in the reasoning. At the end, write a concise conclusion that balances both sides of the discussion. Finally, suggest one policy recommendation that could be derived from the passage. Your answer should be detailed but limited to 300 words. Remember to include citations in APA format whenever possible. Here is the passage that you need to process {passage}.
\end{lstlisting}
\end{tcolorbox}
\label{fig:prompt_segmentation_example}
\end{figure}
Below we present the segmentation of the above prompt template using the three methods described in Section~\ref{ref_subsection_prompt_template_segmentation}.
\begin{enumerate}

    \item \textbf{Pre-defined segmentation}:  
    A pre-defined segmentation method depends on prompt management practices (e.g., storing each segment in a separate text file or Jinja template). For instance, if the role assignment (``You are an expert in climate change and environmental policy'') is designated as one unit and all remaining text as another, the segmentation naturally follows that template choice.
    
    \item \textbf{Structure-aware segmentation}:  
    A sentence-level approach that splits the text into 11 distinct units (one per sentence). For example:
    \begin{itemize}
        \item \textit{1st sentence}: ``You are an expert in climate change and environmental policy.''
        \item \textit{2nd sentence}: ``Please read the following passage carefully.''
        \item \dots
        \item \textit{11th sentence}: ``Here is the passage that you need to process \{passage\}.''
    \end{itemize}

    \item \textbf{LLM-driven segmentation}:  
    A semantic approach that groups content into four logical sections:
    \begin{itemize}
        \item \textit{Role assignment}: ``You are an expert in climate change and environmental policy.''
        \item \textit{Task instructions}: ``Please read the following passage carefully \ldots write a concise conclusion \ldots''
        \item \textit{Format instructions}: ``Your answer should be detailed but limited to 300 words \ldots''
        \item \textit{Input}: ``Here is the passage \ldots \{passage\}.''
    \end{itemize}

\end{enumerate}

\newpage
\section{LLM-driven Attribution Estimation Algorithm}
\begin{algorithm}
\small
\caption{LLM-Driven Prompt Attribution Estimation}
\label{ref_algorithm_llm_driven_attribution_estimation}
\textbf{Input:} LLM; prompt segments $p = [p_1, \ldots, p_M]$; evaluator $s(y, \hat{y})$; dataset $\mathcal{D}_{\text{train}} = \{(x_i, y_i)\}_{i=1}^N$; number of text units to keep $k$; number of experiment $t$; \\
\textbf{Output:} Attribution scores $[a_1, \ldots, a_M]$
\begin{algorithmic}[1]
\State $\{\mathcal{K}^1,\ldots \mathcal{K}^t\} = \text{LLM}(\texttt{AskLLMForIndex}(p, k, t))$ \Comment{prompt LLM to generate $t$ index sets (see Figure~\ref{ref_prompt_for_generating_mask_for_attribution_analysis})}
\Comment{get performance for each mask}
\For{$i = 1$ to $t$}
    \State $s^i = \frac 1N \sum_{j=1}^N s(y_j, \text{LLM}(p_{\mathcal{K}^i}(x_j)))$. 
\EndFor
\State $\pi  = \text{LLM}(\texttt{RankPrompt}(p, \{s^1, \ldots, s^t\}, \{\mathcal{K}^1,\ldots \mathcal {K}^t\} ))$ \Comment{Prompt LLM to rank segments to reflect importance (see Appendix~\ref{ref_prompt_for_estimating_attribution_ranking})}
\For{$j = 1$ to $M$}
    \State $a_j \leftarrow 1 / \pi(j)$ \Comment{LLM gives ranks; alternatively, we can also prompt LLM for scores.}
\EndFor
\State \textbf{return} $[a_1, \ldots, a_M]$
\end{algorithmic}
\end{algorithm}

\newpage
\onecolumn
\section{Initial Prompt Template for RQ1 and RQ2}
\label{ref_appendix_compression_initial_prompt_template}

To ensure that our compression experiments start from representative and high-quality prompts, we constructed an initial prompt template comprising five widely adopted segment types: role-playing, zero-shot chain-of-thought (CoT) prompting, few-shot CoT examples, question placeholders, and context placeholders in Table~\ref{ref_table_prompt_template}. These segments reflect common best practices, with sources cited either from the datasets where they were used or from the original papers that proposed them.

\begin{table*}[hb]
\centering
\scriptsize
\renewcommand{\arraystretch}{1.15}
\setlength{\tabcolsep}{3.5pt}

\begin{tabular}{p{1.5cm} p{2.6cm} p{2.6cm} p{2.6cm} p{2.6cm} p{2.6cm}}
\toprule
\textbf{Component} & \textbf{GSM8K} & \textbf{SQuAD} & \textbf{BBH} & \textbf{HumanEval} & \textbf{MMLU} \\
\midrule

\textbf{Role-playing} &
You are an expert in grade-school math. \cite{mirzadeh2024gsm} &
You are an expert in reading comprehension and QA. &
You are an expert in \texttt{\{bbh\_task\_name\}}. &
You are an expert in Python programming. \cite{zhong2024debug} &
You are an expert in \texttt{\{mmlu\_task\_name\}}. \\

\textbf{Zero-shot CoT} &
Let's think step-by-step before generating the final answer. \newline \texttt{\{gsm8k\_cot\}} \cite{wei2022chain, jiang2023llmlingua, fu2022complexity, fu2023chain} &
Let's think step-by-step before answering the question. \cite{kojima2022large} &
Let's think step-by-step before generating the final answer. \newline \texttt{\{bbh\_cot\}} \cite{suzgun2023challenging} &
Let's think step-by-step before generating the answer. \cite{kojima2022large} &
Let's think step-by-step before generating the final answer. \newline \texttt{\{mmlu\_3shot\_cot\}} \cite{fu2023chain} \\

\textbf{Few-shot CoT Examples} &
Here are a few examples: \texttt{\{examples\}} \cite{mirzadeh2024gsm} &
-- &
Here are a few examples: \texttt{\{examples\}} \cite{brown2020language} &
-- &
Here are a few examples: \texttt{\{examples\}} \\

\textbf{Context} &
-- &
Here is the context you can use to answer the question: \texttt{\{context\}} \cite{li2024500xcompressor, li2024prompt} &
--  &
--  &
--  \\

\textbf{Question Placeholder} &
Here is the question you need to answer. \texttt{\{question\}} \cite{mirzadeh2024gsm, jiang2023llmlingua} &
Here is the question you need to answer. \texttt{\{question\}} \cite{li2024prompt, li2024500xcompressor} &
Here is the question you need to answer. \texttt{\{question\}} \cite{suzgun2023challenging} &
Here is the initial code that you need to complete. \texttt{\{initial\_code\}} \cite{roziere2023code} &
Here is the question you need to answer. \texttt{\{question\}} \\

\bottomrule
\end{tabular}
\caption{Prompt template design across five datasets. Each cell shows the content or placeholder used for a given component. Citations indicate the source of each design choice.}
\label{ref_table_prompt_template}
\end{table*}

\clearpage
\onecolumn
\section{Compressed Prompt Template}
\label{ref_appendix_compressed_prompt_template}
To complement the quantitative results in the main text, we provide qualitative examples of prompt compression across different methods here. Table~\ref{tab:appendix_compressed_prompt} shows the original and compressed prompt templates for the SQuAD dataset under a compression ratio of 50\%.

\begin{table}[ht]
\centering
\scriptsize
\renewcommand{\arraystretch}{1.15}
\setlength{\tabcolsep}{3pt}
\begin{tabular}{
  >{\raggedright\arraybackslash}p{3cm}
  >{\raggedright\arraybackslash}p{9cm}
  >{\raggedright\arraybackslash}p{3cm}}
\toprule
\textbf{Method} & \textbf{Prompt Template} & \textbf{Note} \\
\midrule

Original Template & 
\begin{minipage}[t]{\linewidth}\ttfamily\scriptsize
You are an expert in \{role\_domain\}.\\
Let's think step-by-step before generating the final answer.\\
Here is the context you can use to answer the question:\\
<context>\\
\{context\}\\
</context>\\
Please answer the question below by extracting the minimal span from the context, if available, that best answers the question.\\
<question>\\
\{question\}\\
</question>\\
\end{minipage}
& -- \\
\addlinespace[6pt]

ProCUT (Our Method) &
\begin{minipage}[t]{\linewidth}\ttfamily\scriptsize
Here is the context you can use to answer the question:\\
<context>\\
\{context\}\\
</context>\\
Please answer the question below by extracting the minimal span from the context, if available, that best answers the question.\\
<question>\\
\{question\}\\
</question>\\
\end{minipage}
& Context and question placeholders retained (high attribution). \\
\addlinespace[6pt]

Vanilla LLM (Baseline) &
\begin{minipage}[t]{\linewidth}\ttfamily\scriptsize
Here is the context you can use to answer the question:\\
<context>\\
\{context\}\\
</context>\\
Please answer the question below by extracting the minimal span from the context, if available, that best answers the question.\\
<question>\\
\{question\}\\
</question>\\
\end{minipage}
& Context and question placeholders retained. \\
\addlinespace[6pt]

Selective-Context (Baseline) &
\begin{minipage}[t]{\linewidth}\ttfamily\scriptsize
You are an expert in \{role\_domain\}. Please answer the question below by extracting the minimal span from the context, if available, that best answers the question. <question> \{question\} </question>.
\end{minipage}
& Context placeholder removed, leading to minor drop. \\
\addlinespace[6pt]

LLMLingua (Baseline) &
\begin{minipage}[t]{\linewidth}\ttfamily\scriptsize
You are an in \{role\_domain\}.'s thinkstep before final. Please question below by extracting the minimal span from the context, if available, that best answers the question. <question> \{question\} </question>. \\
\end{minipage}
& Context placeholder removed, leading to minor drop. \\
\addlinespace[6pt]

LLMLingua-2 (Baseline) &
\begin{minipage}[t]{\linewidth}\ttfamily\scriptsize
expert in reading comprehension QA think step-by-step before final answer context to answer question:<context>\{context\}<context> answer question extracting minimal span from context best answers question.\\
\end{minipage}
& Question placeholder removed, leading to large performance drop. \\

\bottomrule
\end{tabular}
\caption{Prompt templates for the SQuAD dataset under compression ratio 50\%.}
\label{tab:appendix_compressed_prompt}
\end{table}

\clearpage
\onecolumn
\section{Prompt Template Compression Performance (Task Level)}
\label{ref_appendix_compression_performance_task_level}

Table~\ref{tab:exp-2-compression-all-41mini} and Table~\ref{tab:exp-2-compression-all-41} report prompt compression performance across 12 tasks from five benchmark datasets under varying compression ratios, using GPT-4.1 mini and GPT-4.1, respectively.

\subsection{GPT-4.1 mini}

\begin{table}[hb]
\centering
\scriptsize
\setlength{\tabcolsep}{3pt}
\begin{tabular}{lcccccccccccc}
\specialrule{1.2pt}{1pt}{1pt}
\textbf{Method} 
& \multicolumn{2}{c}{\textbf{General QA}} 
& \textbf{Coding} 
& \multicolumn{5}{c}{\textbf{BBH Tasks}} 
& \multicolumn{4}{c}{\textbf{MMLU Tasks}} \\
\cmidrule(lr){2-3} \cmidrule(lr){4-4} \cmidrule(lr){5-9} \cmidrule(lr){10-13}
& GSM8K & SQuAD & HumanEval 
& geo. & object & colored & penguins & temporal
& medicine & math & anatomy & astronomy \\
\specialrule{1.2pt}{1pt}{1pt}
\multicolumn{13}{l}{\textbf{Compression Ratio = 0.25 (3/4 components removed)}} \\
\hline
Random Selection (Baseline)& 0.004 & 0.059 & -- & 0.096 & 0.026 & 0.140 & 0.170 & 0.288 & 0.210 & 0.195 & 0.056 & 0.374 \\
Vanilla LLM (Baseline) & 0.132 & 0.059 & -- & 0.000 & 0.000 & 0.124 & 0.000 & 0.000 & 0.162 & 0.000 & 0.000 & 0.000 \\
Selective Context (Baseline) & 0.000 & 0.163 & -- & 0.000 & 0.000 & 0.000 & 0.000 & 0.000 & 0.000 & 0.000 & 0.000 & 0.000 \\
Brute Force (Oracle) & 0.636 & 0.228 & -- & 0.504 & 0.390 & \textbf{0.662} & 0.490 & 0.448 & 0.810 & 0.973 & 0.892 & 0.940 \\
\hline
ProCut (SHAP) & \textbf{0.648} & 0.242 & -- & 0.508 & 0.386 & \textbf{0.662} & 0.474 & 0.474 & 0.822 & 0.981 & 0.896 & \textbf{0.942} \\
ProCut (Leave-One-Out) & 0.636 & 0.246 & -- & 0.502 & \textbf{0.392} & 0.648 & 0.482 & 0.440 & 0.820 & 0.968 & 0.896 & 0.938 \\
ProCut (LASSO) & 0.638 & \textbf{0.247} & -- & \textbf{0.514} & 0.382 & 0.660 & 0.470 & 0.468 & \textbf{0.828} & 0.965 & \textbf{0.900} & \textbf{0.942} \\
ProCut (Greedy Forward) & 0.636 & 0.239 & -- & 0.490 & 0.390 & 0.646 & 0.480 & \textbf{0.472} & 0.816 & \textbf{0.976} & \textbf{0.900} & 0.938 \\
ProCut (LLM-as-Ranker 2 Shots) & 0.636 & 0.242 & -- & 0.506 & 0.368 & 0.654 & 0.486 & 0.468 & 0.818 & 0.965 & 0.894 & 0.940 \\
ProCut (LLM-as-Ranker 4 Shots) & 0.634 & 0.225 & -- & 0.506 & 0.380 & 0.656 & \textbf{0.492} & 0.368 & 0.816 & 0.971 & 0.894 & 0.940 \\
\specialrule{0.8pt}{1pt}{1pt}
\multicolumn{13}{l}{\textbf{Compression Ratio = 0.5 (2/4 components removed)}} \\
\hline
Random Selection (Baseline) & 0.174 & 0.218 & 0.504 & 0.222 & 0.198 & 0.268 & 0.468 & 0.418 & 0.424 & 0.432 & 0.672 & 0.808 \\
Vanilla LLM (Baseline) & 0.582 & 0.750 & 0.000 & 0.550 & 0.914 & 0.728 & 0.580 & 0.446 & 0.824 & 0.976 & 0.918 & \textbf{0.952} \\
Selective Context (Baseline) & 0.006 & 0.235 & 0.000 & 0.000 & 0.068 & 0.106 & 0.210 & 0.060 & 0.264 & 0.267 & 0.284 & 0.302 \\
Brute Force (Oracle) & \textbf{0.856} & 0.744 & 0.834 & 0.558 & \textbf{1.000} & 0.976 & 0.964 & 0.926 & 0.836 & 0.960 & 0.918 & 0.946 \\
\hline
ProCut (SHAP) & 0.848 & 0.746 & 0.832 & \textbf{0.636} & 0.792 & 0.978 & 0.962 & \textbf{0.952} & 0.822 & \textbf{0.971} & 0.922 & 0.940 \\
ProCut (Leave-One-Out) & 0.848 & 0.748 & 0.826 & 0.594 & 0.832 & \textbf{0.984} & 0.960 & 0.942 & 0.822 & 0.965 & 0.914 & 0.940 \\
ProCut (LASSO) & 0.854 & \textbf{0.752} & \textbf{0.842} & 0.622 & 0.826 & \textbf{0.984} & \textbf{0.970} & 0.946 & 0.834 & 0.957 & \textbf{0.924} & 0.944 \\
ProCut (Greedy Forward) & 0.848 & 0.745 & 0.832 & 0.630 & \textbf{1.000} & 0.974 & 0.966 & 0.940 & 0.826 & 0.963 & 0.904 & 0.950 \\
ProCut (LLM-as-Ranker 2 Shots) & 0.846 & 0.746 & 0.824 & 0.608 & 0.776 & 0.974 & 0.966 & 0.922 & \textbf{0.840} & 0.965 & 0.920 & 0.946 \\
ProCut (LLM-as-Ranker 4 Shots) & 0.638 & 0.747 & 0.838 & 0.570 & 0.916 & 0.762 & 0.872 & 0.590 & 0.834 & \textbf{0.971} & 0.914 & 0.948 \\
\specialrule{0.8pt}{1pt}{1pt}
\multicolumn{13}{l}{\textbf{Compression Ratio = 0.75 (1/4 components removed)}} \\
\hline
Random Selection (Baseline)& 0.332 & 0.399 & 0.662 & 0.518 & 0.434 & 0.424 & 0.762 & 0.666 & 0.822 & 0.973 & 0.792 & 0.944 \\
Vanilla LLM (Baseline) & 0.560 & 0.735 & 0.830 & 0.548 & \textbf{1.000} & 0.832 & 0.556 & 0.490 & 0.814 & 0.965 & 0.908 & 0.936 \\
Selective Context (Baseline) & 0.004 & 0.215 & 0.000 & 0.000 & 0.074 & 0.032 & 0.098 & 0.022 & 0.232 & 0.251 & 0.312 & 0.276 \\
Brute Force (Oracle) & 0.814 & 0.739 & 0.840 & 0.602 & 0.990 & 0.984 & 0.950 & 0.798 & 0.824 & 0.963 & 0.920 & 0.944 \\
\hline
ProCut (SHAP) & \textbf{0.830} & 0.737 & 0.830 & \textbf{0.648} & 0.988 & \textbf{0.986} & 0.956 & \textbf{0.886} & 0.828 & \textbf{0.979} & 0.910 & 0.942 \\
ProCut (Leave-One-Out) & 0.804 & 0.739 & 0.832 & 0.616 & 0.988 & \textbf{0.986} & 0.952 & 0.904 & 0.820 & 0.976 & 0.918 & \textbf{0.948} \\
ProCut (LASSO) & 0.798 & 0.731 & 0.832 & 0.616 & 0.988 & 0.984 & \textbf{0.958} & 0.750 & 0.820 & 0.971 & 0.910 & 0.944 \\
ProCut (Greedy Forward) & 0.816 & \textbf{0.740} & 0.832 & 0.614 & \textbf{1.000} & 0.982 & 0.950 & 0.884 & \textbf{0.830} & 0.973 & \textbf{0.920} & \textbf{0.948} \\
ProCut (LLM-as-Ranker 2 Shots) & 0.786 & 0.732 & \textbf{0.844} & 0.592 & 0.988 & 0.982 & 0.954 & 0.740 & 0.824 & 0.973 & 0.918 & 0.946 \\
ProCut (LLM-as-Ranker 4 Shots) & 0.614 & 0.736 & 0.818 & 0.594 & \textbf{1.000} & 0.924 & 0.870 & 0.750 & 0.828 & 0.955 & 0.914 & 0.944 \\
\specialrule{0.8pt}{1pt}{1pt}
\multicolumn{13}{l}{\textbf{No Compression}} \\
\hline
-  & 0.782 &	0.729	&0.837	&0.612	&0.995	&0.993	&0.959	&0.566	&0.819	&0.964	& 0.915	&0.939\\
\specialrule{1.2pt}{1pt}{1pt}
\end{tabular}
\caption{
Performance across tasks at different compression ratios with GPT-4.1 mini. Bolded values indicate the highest score per task at each compression ratio.
}
\label{tab:exp-2-compression-all-41mini}
\end{table}

\newpage
\subsection{GPT-4.1}

\begin{table}[ht]
\centering
\scriptsize
\setlength{\tabcolsep}{3pt}
\begin{tabular}{lcccccccccccc}
\specialrule{1.2pt}{1pt}{1pt}
\textbf{Method} 
& \multicolumn{2}{c}{\textbf{General QA}} 
& \textbf{Coding} 
& \multicolumn{5}{c}{\textbf{BBH Tasks}} 
& \multicolumn{4}{c}{\textbf{MMLU Tasks}} \\
\cmidrule(lr){2-3} \cmidrule(lr){4-4} \cmidrule(lr){5-9} \cmidrule(lr){10-13}
& GSM8K & SQuAD & HumanEval 
& geo. & object & colored & penguins & temporal
& medicine & math & anatomy & astronomy \\
\specialrule{1.2pt}{1pt}{1pt}
\multicolumn{13}{l}{\textbf{Compression Ratio = 0.25 (3/4 components removed)}} \\
\hline
Random Selection (Baseline) & 0.004 & 0.031 & -- & 0.210 & 0.200 & 0.182 & 0.358 & 0.598 & 0.000 & 0.155 & 0.734 & 0.190 \\
  Vanilla LLM (Baseline) & 0.174 & 0.102 & -- & 0.092 & 0.000 & 0.180 & 0.000 & 0.000 & 0.172 & 0.309 & 0.184 & 0.000 \\
Selective Context (Baseline) & 0.000 & 0.215 & -- & 0.000 & 0.000 & 0.000 & 0.000 & 0.000 & 0.000 & 0.000 & 0.000 & 0.000 \\
Brute Force (Oracle) & 0.860 & 0.297 & -- & 0.470 & 0.766 & 0.886 & 0.894 & \textbf{0.998} & 0.850 & 0.773 & 0.924 & 0.946 \\
\hline
ProCUT (SHAP) & 0.868 & 0.298 & -- & 0.458 & 0.776 & 0.888 & 0.900 & 0.996 & \textbf{0.852} & 0.773 & 0.926 & 0.952 \\
ProCUT (Leave-One-Out) & 0.866 & \textbf{0.302} & -- & 0.458 & 0.762 & 0.882 & 0.904 & 0.994 & \textbf{0.852} & 0.768 & 0.922 & 0.950 \\
ProCUT (LASSO) & 0.862 & 0.293 & -- & 0.468 & 0.770 & 0.890 & 0.884 & 0.996 & 0.850 & 0.757 & 0.922 & \textbf{0.954} \\
ProCUT (Greedy Forward) & \textbf{0.870} & 0.300 & -- & 0.464 & 0.776 & 0.884 & 0.892 & 0.996 & 0.850 & 0.781 & 0.926 & 0.948 \\
ProCUT (LLM-ranker 2shot) & 0.866 & 0.099 & -- & \textbf{0.472} & \textbf{0.784} & 0.888 & 0.896 & 0.798 & 0.850 & 0.768 & 0.736 & 0.758 \\
ProCUT (LLM-ranker 4shot) & 0.868 & 0.145 & -- & 0.466 & 0.768 & \textbf{0.890} & 0.886 & 0.996 & 0.680 & \textbf{0.787} & \textbf{0.928} & 0.952 \\

\specialrule{0.8pt}{1pt}{1pt}
\multicolumn{13}{l}{\textbf{Compression Ratio = 0.5 (2/4 components removed)}} \\
\hline
Random Selection (Baseline) & 0.176 & 0.141 & 0.340 & 0.414 & 0.600 & 0.202 & 0.784 & 0.594 & 0.862 & 0.571 & 0.750 & 0.386 \\
Vernilla LLM (Baseline) & 0.846 & \textbf{0.774} & 0.688 & 0.736 & 0.954 & 0.990 & \textbf{1.000} & 0.990 & \textbf{0.870} & 0.928 & 0.948 & 0.964 \\
Selective Context (Baseline) & 0.016 & 0.322 & 0.000 & 0.112 & 0.000 & 0.000 & 0.000 & 0.000 & 0.000 & 0.000 & 0.000 & 0.000 \\
Brute Force (Oracle)& 0.894 & 0.766 & 0.854 & \textbf{0.792} & \textbf{1.000} & 0.996 & \textbf{1.000} & \textbf{0.998} & 0.858 & 0.944 & 0.938 & 0.962 \\
\hline
ProCUT (SHAP) & 0.840 & 0.771 & 0.848 & 0.788 & \textbf{1.000} & 0.992 & \textbf{1.000} & 0.992 & 0.856 & 0.936 & 0.926 & 0.962 \\
ProCUT (Leave-One-Out) & 0.854 & 0.768 & 0.846 & 0.648 & \textbf{1.000} & 0.994 & \textbf{1.000} & 0.994 & 0.860 & \textbf{0.949} & 0.936 & \textbf{0.968} \\
ProCUT (LASSO) & 0.854 & 0.766 & 0.852 & 0.722 & \textbf{1.000} & 0.994 & \textbf{1.000} & 0.994 & 0.846 & 0.875 & 0.940 & 0.962 \\
ProCUT (Greedy Forward) & 0.840 & 0.766 & 0.852 & 0.778 & 0.994 & 0.992 & \textbf{1.000} & 0.990 & \textbf{0.870} & 0.917 & \textbf{0.952} & 0.960 \\
ProCUT (LLM-ranker 2shot) & \textbf{0.892} & 0.763 & \textbf{0.862} & 0.730 & 0.998 & 0.992 & \textbf{1.000} & 0.792 & 0.852 & 0.917 & 0.938 & 0.966 \\
ProCUT (LLM-ranker 4shot) & 0.890 & 0.764 & 0.858 & 0.722 & \textbf{1.000} & 0.984 & \textbf{1.000} & 0.994 & 0.850 & 0.947 & 0.936 & \textbf{0.968} \\

\specialrule{0.8pt}{1pt}{1pt}
\multicolumn{13}{l}{\textbf{Compression Ratio = 0.75 (1/4 components removed)}} \\
\hline
Random Selection (Baseline) & 0.534 & 0.383 & 0.848 & 0.796 & 0.600 & 0.396 & 0.800 & 0.594 & \textbf{0.854} & 0.757 & 0.750 & 0.768 \\
Vernilla LLM (Baseline) & 0.842 & 0.772 & 0.850 & 0.826 & 0.994 & 0.994 & \textbf{1.000} & 0.990 & 0.848 & 0.949 & \textbf{0.950} & 0.964 \\
Selective Context (Baseline) & 0.012 & 0.272 & 0.000 & 0.034 & 0.044 & 0.000 & 0.000 & 0.000 & 0.000 & 0.000 & 0.000 & 0.000 \\
Brute Force (Oracle) & 0.886 & 0.769 & 0.852 & 0.800 & \textbf{1.000} & 0.992 & \textbf{1.000} & 0.994 & 0.852 & 0.944 & 0.930 & 0.964 \\
\hline
ProCUT (SHAP) & 0.850 & 0.774 & \textbf{0.858} & 0.800 & \textbf{1.000} & 0.994 & \textbf{1.000} & 0.994 & 0.846 & 0.941 & 0.926 & \textbf{0.966} \\
ProCUT (Leave-One-Out) & 0.874 & 0.764 & \textbf{0.858} & 0.776 & \textbf{1.000} & 0.992 & \textbf{1.000} & 0.990 & 0.840 & 0.949 & 0.928 & 0.960 \\
ProCUT (LASSO) & 0.882 & 0.765 & \textbf{0.858} & 0.798 & \textbf{1.000} & 0.996 & \textbf{1.000} & 0.992 & 0.852 & 0.955 & 0.932 & 0.964 \\
ProCUT (Greedy Forward) & 0.868 & 0.775 & 0.850 & 0.802 & 0.998 & \textbf{1.000} & \textbf{1.000} & 0.990 & \textbf{0.854} & 0.949 & 0.944 & 0.958 \\
ProCUT (LLM-ranker 2shot) & \textbf{0.898} & 0.774 & 0.852 & \textbf{0.808} & \textbf{1.000} & 0.994 & \textbf{1.000} & \textbf{0.996} & 0.850 & 0.947 & 0.936 & \textbf{0.966} \\
ProCUT (LLM-ranker 4shot) & 0.896 & \textbf{0.790} & 0.850 & 0.770 & \textbf{1.000} & 0.996 & \textbf{1.000} & 0.992 & 0.852 & \textbf{0.960} & 0.932 & 0.964 \\

\specialrule{0.8pt}{1pt}{1pt}
\multicolumn{13}{l}{\textbf{No Compression}} \\
\hline
-  & 0.890	& 0.795	& 0.865	& 0.791	& 1.000	& 0.993	& 1.000	& 0.995	& 0.849	& 0.947	& 0.936	& 0.963\\
\specialrule{1.2pt}{1pt}{1pt}
\end{tabular}
\caption{
Performance across tasks at different compression ratios with GPT-4.1 model. Bolded values indicate the highest score per task at each compression ratio.
}
\label{tab:exp-2-compression-all-41}
\end{table}

\clearpage
\onecolumn
\section{Prompt Template Compression Performance (Dataset Level)}
\label{ref_appendix_compression_performance_dataset_level}

Table~\ref{tab:compression-performance-aggregated-gpt-41-mini} and Table~\ref{tab:compression-performance-aggregated-gpt-41} summarize the detailed prompt compression performance across five benchmark datasets under varying compression ratios, for GPT-4.1 mini and GPT-4.1 respectively.

\subsection{GPT-4.1 mini}

\begin{table}[ht]
\centering
\small
\setlength{\tabcolsep}{5pt}
\begin{tabular}{lcccccc}
\toprule
\textbf{Method} & \textbf{GSM8K} & \textbf{SQuAD} & \textbf{HumanEval} & \textbf{BBH} & \textbf{MMLU} & \textbf{Average} \\
\midrule
\multicolumn{7}{l}{\textbf{Compression Ratio $\lambda$ = 0.25 (3/4 components removed)}} \\
\midrule
Random Selection (Baseline)& 0.004 & 0.059 & -- & 0.144 & 0.209 & 0.104 \\
Vanilla LLM (Baseline) & 0.132 & 0.059 & -- & 0.025 & 0.041 & 0.064 \\
Selective Context (Baseline) & 0.000 & 0.163 & -- & 0.000 & 0.000 & 0.041 \\
Brute Force (Oracle) & 0.636 & 0.228 & --   & 0.499 & 0.904 & 0.567 \\
\hline
ProCut (SHAP) & \textbf{0.648} & 0.242 & -- & \textbf{0.501} & \textbf{0.910} & \textbf{0.575} \\
ProCut (Leave-One-Out) & 0.636 & 0.246 & -- & 0.493 & 0.906 & 0.570 \\
ProCut (LASSO) & 0.638 & \textbf{0.247} & -- & 0.499 & 0.909 & 0.573 \\
ProCut (Greedy Forward) & 0.636 & 0.239 & -- & 0.496 & 0.908 & 0.569 \\
ProCut (LLM-as-Ranker 2 Shots) & 0.636 & 0.242 & -- & 0.496 & 0.904 & 0.570 \\
ProCut (LLM-as-Ranker 4 Shots) & 0.634 & 0.225 & -- & 0.480 & 0.905 & 0.561 \\
\midrule
\multicolumn{7}{l}{\textbf{Compression Ratio $\lambda$ = 0.5 (2/4 components removed)}} \\
\midrule
Random Selection (Baseline) & 0.174 & 0.218 & 0.504 & 0.315 & 0.584 & 0.359 \\
Vanilla LLM (Baseline) & 0.582 & 0.750 & 0.000 & 0.644 & 0.918 & 0.579 \\
Selective Context (Baseline) & 0.006 & 0.235 & 0.000 & 0.089 & 0.279 & 0.122 \\
Brute Force (Oracle) & \textbf{0.856} & 0.744 & 0.834 & 0.885 & 0.915 & 0.847 \\
\hline
ProCut (SHAP) & 0.848 & 0.746 & 0.832 & 0.864 & 0.914 & 0.841 \\
ProCut (Leave-One-Out) & 0.848 & 0.748 & 0.826 & 0.862 & 0.910 & 0.839 \\
ProCut (LASSO) & 0.854 & \textbf{0.752} & \textbf{0.842} & 0.870 & 0.915 & 0.846 \\
ProCut (Greedy Forward) & 0.848 & 0.745 & 0.832 & \textbf{0.902} & 0.911 & \textbf{0.848} \\
ProCut (LLM-as-Ranker 2 Shots) & 0.846 & 0.746 & 0.824 & 0.849 & \textbf{0.918} & 0.837 \\
ProCut (LLM-as-Ranker 4 Shots) & 0.638 & 0.747 & 0.838 & 0.742 & 0.917 & 0.776 \\
\midrule
\multicolumn{7}{l}{\textbf{Compression Ratio $\lambda$ = 0.75 (1/4 components removed)}} \\
\midrule
Random Selection (Baseline) & 0.332 & 0.399 & 0.662 & 0.561 & 0.883 & 0.567 \\
Vanilla LLM (Baseline) & 0.560 & 0.735 & 0.830 & 0.685 & 0.906 & 0.743 \\
Selective Context (Baseline) & 0.004 & 0.215 & 0.000 & 0.045 & 0.268 & 0.106 \\
Brute Force (Oracle) & 0.814 & 0.739 & 0.840 & 0.865 & 0.913 & 0.834 \\
\hline
ProCut (SHAP) & \textbf{0.830} & 0.737 & 0.830 & \textbf{0.893} & 0.915 & \textbf{0.841} \\
ProCut (Leave-One-Out) & 0.804 & 0.739 & 0.832 & 0.889 & 0.916 & 0.836 \\
ProCut (LASSO) & 0.798 & 0.731 & 0.832 & 0.859 & 0.911 & 0.826 \\
ProCut (Greedy Forward) & 0.816 & \textbf{0.740} & 0.832 & 0.886 & \textbf{0.918} & 0.838 \\
ProCut (LLM-as-Ranker 2 Shots) & 0.786 & 0.732 & \textbf{0.844} & 0.851 & 0.915 & 0.826 \\
ProCut (LLM-as-Ranker 4 Shots) & 0.614 & 0.736 & 0.818 & 0.828 & 0.910 & 0.781 \\
\specialrule{0.8pt}{1pt}{1pt}
\multicolumn{7}{l}{\textbf{No Compression}} \\
\hline
- & 0.782 & 0.729 & 0.837 & 0.826 & 0.909 & 0.817 \\
\bottomrule
\end{tabular}
\caption{Performance with GPT-4.1 mini across tasks at different compression ratios. Bolded values indicate the highest score per dataset at each compression ratio.}
\label{tab:compression-performance-aggregated-gpt-41-mini}
\end{table}

\newpage
\subsection{GPT-4.1}

\begin{table}[ht]
\centering
\small
\setlength{\tabcolsep}{5pt}
\begin{tabular}{lcccccc}
\toprule
\textbf{Method} & \textbf{GSM8K} & \textbf{SQuAD} & \textbf{HumanEval} & \textbf{BBH} & \textbf{MMLU} & \textbf{Average} \\
\midrule
\multicolumn{7}{l}{\textbf{Compression Ratio $\lambda$ = 0.25 (3/4 components removed)}} \\
\midrule
Random Selection & 0.004 & 0.031 & --    & 0.310 & 0.270 & 0.154 \\
Vernilla LLM (Baseline) & 0.174 & 0.102 & --    & 0.054 & 0.166 & 0.124 \\
Selective Context (Baseline) & 0.000 & 0.215 & --    & 0.000 & 0.000 & 0.054 \\
Brute Force (Baseline) & 0.860 & 0.297 & --    & 0.803 & 0.873 & 0.708 \\
\hline
ProCUT (SHAP) & 0.868 & 0.298 & --    & \textbf{0.804} & \textbf{0.876} & 0.711 \\
ProCUT (Leave-One-Out) & 0.866 & \textbf{0.302} & --    & 0.800 & 0.873 & 0.710 \\
ProCUT (LASSO) & 0.862 & 0.293 & --    & 0.802 & 0.871 & 0.707 \\
ProCUT (Greedy Forward) & \textbf{0.870} & 0.300 & --    & 0.802 & \textbf{0.876} & \textbf{0.712} \\
ProCUT (LLM-as-ranker 2 shot) & 0.866 & 0.099 & --    & 0.768 & 0.778 & 0.628 \\
ProCUT (LLM-as-ranker 4 shot) & 0.868 & 0.145 & --    & 0.801 & 0.837 & 0.663 \\
\midrule
\multicolumn{7}{l}{\textbf{Compression Ratio $\lambda$ = 0.5 (2/4 components removed)}} \\
\midrule
Random Selection & 0.176 & 0.141 & 0.340 & 0.519 & 0.642 & 0.364 \\
Vernilla LLM (Baseline) & 0.846 & \textbf{0.774} & 0.688 & 0.934 & \textbf{0.928} & 0.834 \\
Selective Context (Baseline) & 0.016 & 0.322 & 0.000 & 0.022 & 0.000 & 0.072 \\
Brute Force (Baseline) & \textbf{0.894} & 0.766 & 0.854 & \textbf{0.957} & 0.926 & \textbf{0.879} \\
\hline
ProCUT (SHAP) & 0.840 & 0.771 & 0.848 & 0.954 & 0.920 & 0.867 \\
ProCUT (Leave-One-Out) & 0.854 & 0.768 & 0.846 & 0.927 & \textbf{0.928} & 0.865 \\
ProCUT (LASSO) & 0.854 & 0.766 & 0.852 & 0.942 & 0.906 & 0.864 \\
ProCUT (Greedy Forward) & 0.840 & 0.766 & 0.852 & 0.951 & 0.925 & 0.867 \\
ProCUT (LLM-as-ranker 2 shot) & 0.892 & 0.763 & \textbf{0.862} & 0.902 & 0.918 & 0.868 \\
ProCUT (LLM-as-ranker 4 shot) & 0.890 & 0.764 & 0.858 & 0.940 & 0.925 & 0.875 \\
\midrule
\multicolumn{7}{l}{\textbf{Compression Ratio $\lambda$ = 0.75 (1/4 components removed)}} \\
\midrule
Random Selection & 0.534 & 0.383 & 0.848 & 0.637 & 0.782 & 0.637 \\
Vernilla LLM (Baseline) & 0.842 & 0.772 & 0.850 & \textbf{0.961} & \textbf{0.928} & 0.871 \\
Selective Context (Baseline) & 0.012 & 0.272 & 0.000 & 0.016 & 0.000 & 0.060 \\
Brute Force (Oracle) & 0.886 & 0.769 & 0.852 & 0.957 & 0.923 & 0.877 \\
\hline
ProCUT (SHAP) & 0.850 & 0.774 & \textbf{0.858} & 0.958 & 0.920 & 0.872 \\
ProCUT (Leave-One-Out) & 0.874 & 0.764 & \textbf{0.858} & 0.952 & 0.919 & 0.873 \\
ProCUT (LASSO) & 0.882 & 0.765 & \textbf{0.858} & 0.957 & 0.926 & 0.878 \\
ProCUT (Greedy Forward) & 0.868 & 0.775 & 0.850 & 0.958 & 0.926 & 0.875 \\
ProCUT (LLM-as-ranker 2 shot) & \textbf{0.898} & 0.774 & 0.852 & 0.960 & 0.925 & 0.882 \\
ProCUT (LLM-as-ranker 4 shot) & 0.896 & \textbf{0.790} & 0.850 & 0.952 & 0.927 & \textbf{0.883} \\
\specialrule{0.8pt}{1pt}{1pt}
\multicolumn{7}{l}{\textbf{No Compression}} \\
\hline
- & 0.890	& 0.795	& 0.865	& 0.956	& 0.924	& 0.886\\
\bottomrule
\end{tabular}
\caption{Performance  with GPT-4.1 across tasks at different compression ratios. Bolded values indicate the highest score per dataset at each compression ratio.}
\label{tab:compression-performance-aggregated-gpt-41}
\end{table}

\onecolumn
\clearpage

\section{Prompt Template Compression Performance (Aggregated)}
\label{ref_appendix_compression_performance_aggregated}
Table~\ref{tab_average_compression_performance_compression_ratio_level_gpt_41mini} and Table~\ref{tab_average_compression_performance_compression_ratio_level_gpt_41} report aggregated prompt compression performance under varying compression ratios, using GPT-4.1 mini and GPT-4.1, respectively.

\subsection{GPT-4.1 mini}
\begin{table}[ht]
\centering
\small                                    
\setlength{\tabcolsep}{4pt}                    
\begin{tabular}{lcccc}
\specialrule{1.2pt}{1pt}{1pt}

  & \multicolumn{3}{c}{\textbf{Compression Ratio}} 
  &  \\
\cmidrule(lr){2-4}
\textbf{Method} & \textbf{25\%} & \textbf{50\%} & \textbf{75\%} & \textbf{Average} \\      
\specialrule{1.2pt}{1pt}{1pt}
Random Selection (Baseline) & 0.104 & 0.359 & 0.567 & 0.343 \\
Vanilla LLM (Baseline) & 0.064 & 0.579 & 0.743 & 0.462 \\
Selective Context (Baseline) & 0.041 & 0.122 & 0.106 & 0.090 \\
\midrule
Brute Force (Oracle) & 0.567 & 0.847 & 0.834 & 0.749 \\
\midrule
ProCut (SHAP) & \textbf{0.575} & 0.841 & \textbf{0.841} & \textbf{0.752} \\
ProCut (Leave-One-Out) & 0.570 & 0.839 & 0.836 & 0.748 \\
ProCut (LASSO) & 0.573 & 0.846 & 0.826 & 0.748 \\
ProCut (Greedy Forward) & 0.569 & \textbf{0.848} & 0.838 & \textbf{0.752} \\
ProCut (LLM-as-Ranker 2 Shots) & 0.570 & 0.837 & 0.826 & 0.744 \\
ProCut (LLM-as-Ranker 4 Shots) & 0.561 & 0.776 & 0.781 & 0.706 \\

\specialrule{1.2pt}{1pt}{1pt}
\end{tabular}
\caption{Average performance of compressed prompts with GPT-4.1 mini under different compression ratios. The no-compression baseline achieves an average performance of 0.817.}
\label{tab_average_compression_performance_compression_ratio_level_gpt_41mini}
\end{table}

\subsection{GPT-4.1}
\begin{table}[ht]
\centering
\small                                    
\setlength{\tabcolsep}{4pt}                    
\begin{tabular}{lcccc}
\specialrule{1.2pt}{1pt}{1pt}

  & \multicolumn{3}{c}{\textbf{Compression Ratio}} 
  &  \\
\cmidrule(lr){2-4}
\textbf{Method} & \textbf{25\%} & \textbf{50\%} & \textbf{75\%} & \textbf{Average} \\      
\specialrule{1.2pt}{1pt}{1pt}
Random Selection & 0.154 & 0.364 & 0.637 & 0.385 \\
Vanilla LLM (Baseline) & 0.124 & 0.834 & 0.871 & 0.610 \\
Selective Context (Baseline) & 0.054 & 0.072 & 0.060 & 0.062 \\
\midrule
Brute Force (Oracle) & 0.708 & \textbf{0.879} & 0.877 & \textbf{0.822} \\
\midrule
ProCUT (SHAP) & 0.711 & 0.867 & 0.872 & 0.817 \\
ProCUT (Leave-One-Out) & 0.710 & 0.865 & 0.873 & 0.816 \\
ProCUT (LASSO) & 0.707 & 0.864 & 0.878 & 0.816 \\
ProCUT (Greedy Forward) & \textbf{0.712} & 0.867 & 0.875 & 0.818 \\
ProCUT (LLM-as-ranker 2 shot) & 0.628 & 0.868 & 0.882 & 0.792 \\
ProCUT (LLM-as-ranker 4 shot) & 0.663 & 0.875 & \textbf{0.883} & 0.807 \\

\specialrule{1.2pt}{1pt}{1pt}
\end{tabular}
\caption{Average performance of compressed prompts with GPT-4.1 under different compression ratios. The no-compression baseline achieves an average performance of 0.886.}
\label{tab_average_compression_performance_compression_ratio_level_gpt_41}
\end{table}

\clearpage
\onecolumn
\section{Prompt Template Segments Attribution (Task Level)}

Table~\ref{ref_table_exp_2_attribution_results} presents the attribution of prompt template segments to task performance across 12 tasks from five benchmark datasets.

\begin{table*}[ht]
\centering
\scriptsize
\setlength{\tabcolsep}{3pt}
\begin{tabular}{lcccccccccccc}
\specialrule{1.2pt}{1pt}{1pt}
\textbf{Method} & \multicolumn{2}{c}{\textbf{General QA}} & \textbf{Coding} & \multicolumn{5}{c}{\textbf{BBH Tasks}} & \multicolumn{4}{c}{\textbf{MMLU Tasks}} \\
\cmidrule(lr){2-3} \cmidrule(lr){4-4} \cmidrule(lr){5-9} \cmidrule(lr){10-13}
& GSM8K & SQuAD & HumanEval & geo. & object & colored & penguins & temporal & medicine & math & anatomy & astronomy \\
\specialrule{1.2pt}{1pt}{1pt}
\multicolumn{13}{l}{\textbf{LLM = GPT-4o mini}} \\
\hline
Role-playing & \cellcolor[HTML]{FFFDFD} 0.018 & \cellcolor[HTML]{F3F7FB} -0.014 & \cellcolor[HTML]{F9FBFD} -0.007 & \cellcolor[HTML]{FFFEFE} 0.008 & \cellcolor[HTML]{FFFDFD} 0.015 & \cellcolor[HTML]{FDFDFE} -0.003 & \cellcolor[HTML]{FFFDFD} 0.012 & \cellcolor[HTML]{FFFFFF} 0.003 & \cellcolor[HTML]{CEDCEE} -0.059 & \cellcolor[HTML]{D3E0F0} -0.053 & \cellcolor[HTML]{E0E9F4} -0.038 & \cellcolor[HTML]{D2DFF0} -0.054 \\
Chain-of-thought & \cellcolor[HTML]{FFFFFF} 0.002 & \cellcolor[HTML]{F9FBFD} -0.007 & \cellcolor[HTML]{FFFFFF} 0.003 & \cellcolor[HTML]{FFFBFB} 0.035 & \cellcolor[HTML]{FFFCFC} 0.022 & \cellcolor[HTML]{F1F5FA} -0.017 & \cellcolor[HTML]{FDFEFE} -0.002 & \cellcolor[HTML]{FFFDFD} 0.012 & \cellcolor[HTML]{E4ECF6} -0.033 & \cellcolor[HTML]{C7D7EC} -0.068 & \cellcolor[HTML]{F2F6FA} -0.016 & \cellcolor[HTML]{D3E0F0} -0.053 \\
In-context Learning & \cellcolor[HTML]{FFFCFC} 0.023 & -- & -- & \cellcolor[HTML]{FFEFEF} \textbf{0.125} & \cellcolor[HTML]{FFFDFD} 0.012 & \cellcolor[HTML]{FFDEDE} \textbf{0.260} & \cellcolor[HTML]{FFDDDD} \textbf{0.263} & \cellcolor[HTML]{FFE9E9} \textbf{0.175} & \cellcolor[HTML]{FFF6F6} 0.074 & \cellcolor[HTML]{FFECEC} \textbf{0.146} & \cellcolor[HTML]{FFF4F4} 0.084 & \cellcolor[HTML]{F8FAFC} -0.009 \\
Context Placeholder & -- & \cellcolor[HTML]{FFDCDC} \textbf{0.271} & -- & -- & -- & -- & -- & -- & -- & -- & -- & -- \\
Question Placeholder & \cellcolor[HTML]{FF9494} \textbf{0.837} & \cellcolor[HTML]{FFB7B7} \textbf{0.566} & \cellcolor[HTML]{FF8080} \textbf{0.993} & \cellcolor[HTML]{FFC8C8} \textbf{0.432} & \cellcolor[HTML]{FF9D9D} \textbf{0.772} & \cellcolor[HTML]{FFA4A4} \textbf{0.710} & \cellcolor[HTML]{FFABAB} \textbf{0.657} & \cellcolor[HTML]{FF9898} \textbf{0.81} & \cellcolor[HTML]{FFA6A6} \textbf{0.698} & \cellcolor[HTML]{FFCBCB} \textbf{0.404} & \cellcolor[HTML]{FFB1B1} \textbf{0.609} & \cellcolor[HTML]{FF9191} \textbf{0.866} \\
\specialrule{0.8pt}{1pt}{1pt}
\multicolumn{13}{l}{\textbf{LLM = GPT-4.1 mini}} \\
\hline
Role-playing & \cellcolor[HTML]{FFFEFE} 0.009 & \cellcolor[HTML]{FFFEFE} 0.011 & \cellcolor[HTML]{FFFFFF} 0.000 & \cellcolor[HTML]{F7F9FC} -0.01 & \cellcolor[HTML]{FFFEFE} 0.01 & \cellcolor[HTML]{FFFDFD} 0.018 & \cellcolor[HTML]{EAF0F8} -0.025 & \cellcolor[HTML]{B6CCE6} -0.088 & \cellcolor[HTML]{D8E4F2} -0.047 & \cellcolor[HTML]{ECF2F8} -0.023 & \cellcolor[HTML]{E0E9F4} -0.038 & \cellcolor[HTML]{F2F6FA} -0.016 \\
Chain-of-thought & \cellcolor[HTML]{F0F4FA} -0.018 & \cellcolor[HTML]{FDFEFE} -0.002 & \cellcolor[HTML]{FFFFFF} 0.000 & \cellcolor[HTML]{FFFCFC} 0.023 & \cellcolor[HTML]{FFF2F2} \textbf{0.103} & \cellcolor[HTML]{FFFBFB} 0.033 & \cellcolor[HTML]{FFFCFC} 0.023 & \cellcolor[HTML]{A2BDDF} \textbf{-0.113} & \cellcolor[HTML]{E2EBF5} -0.035 & \cellcolor[HTML]{E5ECF6} -0.032 & \cellcolor[HTML]{D5E1F0} -0.051 & \cellcolor[HTML]{EBF1F8} -0.024 \\
In-context Learning & \cellcolor[HTML]{FFE6E6} \textbf{0.199} & -- & -- & \cellcolor[HTML]{FFF5F5} 0.075 & \cellcolor[HTML]{FFECEC} \textbf{0.147} & \cellcolor[HTML]{FFEAEA} \textbf{0.161} & \cellcolor[HTML]{FFE4E4} \textbf{0.208} & \cellcolor[HTML]{FFE6E6} \textbf{0.198} & \cellcolor[HTML]{FFF5F5} 0.077 & \cellcolor[HTML]{FFF5F5} 0.075 & \cellcolor[HTML]{FFF5F5} 0.079 & \cellcolor[HTML]{FFF4F4} 0.086 \\
Context Placeholder & -- & \cellcolor[HTML]{FFE1E1} \textbf{0.232} & -- & -- & -- & -- & -- & -- & -- & -- & -- & -- \\
Question Placeholder & \cellcolor[HTML]{FF9A9A} \textbf{0.789} & \cellcolor[HTML]{FFBDBD} \textbf{0.519} & \cellcolor[HTML]{FF8080} \textbf{1.000} & \cellcolor[HTML]{FFC4C4} \textbf{0.462} & \cellcolor[HTML]{FFA3A3} \textbf{0.72} & \cellcolor[HTML]{FF9A9A} \textbf{0.789} & \cellcolor[HTML]{FF9C9C} \textbf{0.773} & \cellcolor[HTML]{FFC7C7} \textbf{0.443} & \cellcolor[HTML]{FF9393} \textbf{0.845} & \cellcolor[HTML]{FF9E9E} \textbf{0.76} & \cellcolor[HTML]{FFA1A1} \textbf{0.739} & \cellcolor[HTML]{FF9E9E} \textbf{0.764} \\
\specialrule{0.8pt}{1pt}{1pt}
\multicolumn{13}{l}{\textbf{LLM = GPT-4.1 nano}} \\
\hline
Role-playing & \cellcolor[HTML]{FFF4F4} 0.089 & \cellcolor[HTML]{FFFDFD} 0.015 & \cellcolor[HTML]{EAF0F8} -0.025 & \cellcolor[HTML]{FAFBFD} -0.006 & \cellcolor[HTML]{F7F9FC} -0.01 & \cellcolor[HTML]{FDFDFE} -0.003 & \cellcolor[HTML]{FDFEFE} -0.002 & \cellcolor[HTML]{FFFFFF} 0.002 & \cellcolor[HTML]{D3E0F0} -0.053 & \cellcolor[HTML]{CCDBED} -0.062 & \cellcolor[HTML]{E6EDF6} -0.030 & \cellcolor[HTML]{C0D3E9} -0.076 \\
Chain-of-thought & \cellcolor[HTML]{FFFDFD} 0.016 & \cellcolor[HTML]{FDFDFE} -0.003 & \cellcolor[HTML]{EAF0F8} -0.025 & \cellcolor[HTML]{F4F7FB} -0.013 & \cellcolor[HTML]{FFF8F8} 0.057 & \cellcolor[HTML]{DBE5F2} -0.044 & \cellcolor[HTML]{FFFEFE} 0.007 & \cellcolor[HTML]{FDFDFE} -0.003 & \cellcolor[HTML]{E0E9F4} -0.038 & \cellcolor[HTML]{C8D8EC} -0.067 & \cellcolor[HTML]{EDF2F9} -0.022 & \cellcolor[HTML]{ECF2F8} -0.023 \\
In-context Learning & \cellcolor[HTML]{BACEE7} -0.084 & -- & -- & \cellcolor[HTML]{FFF6F6} 0.071 & \cellcolor[HTML]{FFF4F4} 0.085 & \cellcolor[HTML]{FFF7F7} 0.064 & \cellcolor[HTML]{FFF7F7} 0.060 & \cellcolor[HTML]{FFFEFE} 0.005 & \cellcolor[HTML]{BBCEE7} -0.083 & \cellcolor[HTML]{FFFCFC} 0.025 & \cellcolor[HTML]{FDFEFE} -0.002 & \cellcolor[HTML]{82A7D4} \textbf{-0.151} \\
Context Placeholder & -- & \cellcolor[HTML]{FFE1E1} \textbf{0.237} & -- & -- & -- & -- & -- & -- & -- & -- & -- & -- \\
Question Placeholder & \cellcolor[HTML]{FFDDDD} \textbf{0.269} & \cellcolor[HTML]{FFC9C9} \textbf{0.426} & \cellcolor[HTML]{FF8787} \textbf{0.94} & \cellcolor[HTML]{FFDFDF} \textbf{0.248} & \cellcolor[HTML]{FF9797} \textbf{0.818} & \cellcolor[HTML]{FFBFBF} \textbf{0.503} & \cellcolor[HTML]{FFEEEE} \textbf{0.135} & \cellcolor[HTML]{FFFEFE} 0.007 & \cellcolor[HTML]{FF9B9B} \textbf{0.783} & \cellcolor[HTML]{FFA3A3} \textbf{0.723} & \cellcolor[HTML]{FFA2A2} \textbf{0.733} & \cellcolor[HTML]{FF9C9C} \textbf{0.779} \\
\specialrule{1.2pt}{1pt}{1pt}
\end{tabular}
\caption{Segment‐level attribution of the five prompt components across the 12 tasks (mean of five runs). Cells with $\lvert\text{attribution}\rvert>0.1$ are bolded; darker red shades mark stronger positive contributions, while darker blue shades mark stronger negative contributions.}

\label{ref_table_exp_2_attribution_results}
\end{table*}

\clearpage
\onecolumn
\section{TextGrad and ProCut Prompts}
\label{ref_appendix_textgrad_and_procut_prompt_examples}
In this section, we present illustrative prompts generated after three optimization iterations under two settings: (i) TextGrad-only (Figure~\ref{ref_prompt_for_textgrad_only_prompt_example} and~\ref{ref_prompt_for_textgrad_only_prompt_example_2}) and (ii) TextGrad with ProCut compression at compression ratio $r = 0.2$ (Figure~\ref{ref_prompt_for_textgrad_and_procut_prompt_example}). Both runs were conducted with identical hyperparameter settings on the dataset described in Table~\ref{ref_table_dataset_and_metric_summary}.

\begin{figure}[H]
\begin{tcolorbox}[
    enhanced,
    breakable,
    sharp corners=south,
    title=Prompt from TextGrad only (page 1/2),
    fonttitle=\bfseries,
    colback=white,
    colframe=black,
    left=1pt,
    right=1pt,
    top=1pt,
    bottom=1pt,
    boxsep=1pt
]

\begin{lstlisting}[
    breaklines=true,
    basicstyle=\ttfamily\tiny,
    aboveskip=2pt,
    belowskip=2pt
]
"You are tasked with performing an extractive question answering (QA) task. Given a question and a context, your goal is to extract the exact answer phrase from the context that directly answers the question. Please follow these instructions carefully and exactly:

**Output Format Requirements (Strictly Enforced):**  
- Your entire output must be exactly: `<answer>ExactAnswerText</answer>` with no extra characters, spaces, line breaks, tabs, or invisible characters before, inside, or after the tags.  
- The tags must be exactly lowercase and spelled as `<answer>` and `</answer>`, with no variation, no spaces inside the tag brackets, and no self-closing or alternative forms.  
- Do not include any leading or trailing spaces inside the tags. For example, `<answer> answer </answer>` is invalid.  
- Use only standard ASCII space characters (U+0020) if spaces are part of the answer text. Do not use tabs, non-breaking spaces, zero-width spaces, or any other invisible or special whitespace characters anywhere inside or around the tags or answer text.  
- Output nothing else beyond the tagged answer\u2014no commentary, explanations, line breaks, or formatting.

**Answer Extraction Rules:**  
1. Extract the answer verbatim from the provided context without adding, omitting, or paraphrasing any words. Preserve all original capitalization, spacing, punctuation, Unicode characters (including diacritics, macrons, accents), and special characters exactly as they appear. Do not normalize or alter Unicode characters or punctuation in any way.  
2. Provide the shortest possible contiguous text span from the context that fully answers the question. This minimal answer must include all words (including articles such as \u201ca,\u201d \u201can,\u201d \u201cthe\u201d and other function words) if they are part of the minimal contiguous substring exactly as it appears in the context.  
   - Minimal answer = shortest contiguous substring from the context that fully answers the question, including all words in that substring exactly as they appear.  
   - If the minimal answer includes a leading article such as \u201cthe,\u201d include it exactly as is.  
3. If the question asks for a person\u2019s name, title, or rank, extract only the minimal proper noun phrase (e.g., the person\u2019s name) as it appears in the context, excluding any preceding titles, ranks, or roles unless explicitly required by the question.  
4. For numeric answers (counts, quantities, dates, ranks), extract only the numeric characters exactly as they appear in the context, excluding any accompanying units, descriptors, or qualifiers unless explicitly requested by the question.  
   - Include leading numeric qualifiers such as \u201cat least,\u201d \u201capproximately,\u201d or \u201cabout\u201d if they appear contiguous with the number and are essential to the numeric meaning.
   - Do not include trailing or preceding units, descriptors, or qualifiers such as \u201cdaily readers,\u201d \u201cpeople,\u201d \u201ckilograms,\u201d or \u201cyears\u201d unless explicitly requested.
   - For example:  
     - Question: How many Grammy nominations?  
     - Correct answer: `<answer>10</answer>`  
     - Incorrect answer: `<answer>10 Grammy nominations</answer>`  
     - Question: How big is the audience?  
     - Correct answer: `<answer>at least 30,000</answer>`  
     - Incorrect answer: `<answer>at least 30,000 daily readers</answer>`  

5. Do not include any examples, elaborations, clarifications, or additional descriptive phrases beyond the minimal answer phrase, even if they appear in the context. Specifically:  
   - Exclude any phrases introduced by \u201csuch as,\u201d \u201cfor example,\u201d \u201cincluding,\u201d or similar expressions, even if they appear contiguous with the minimal answer span.  
   - The minimal answer must exclude all such expansions or illustrative content unless explicitly requested by the question.  
   - Avoid including any potentially misspelled, rare, or suspicious terms that appear only as part of examples or elaborations.  
6. Always extract the exact substring from the context without paraphrasing, synonym substitution, abbreviation expansion, or normalization.  
7. If the question is ambiguous or no valid answer can be found, output `<answer></answer>` exactly with no spaces inside the tags.

**Answer Selection and Minimality:**  
8. If multiple valid answers exist, select the one that best matches the expected answer format and is the shortest span that fully answers the question.  
9. When multiple minimal answers are valid, always select the earliest occurrence in the context.  
10. To ensure minimality, first locate the full contiguous text span that answers the question, then iteratively remove any leading or trailing words one at a time that do not affect the completeness of the answer, stopping when further removal would make the answer incomplete.  
11. Include all articles and function words that are part of the minimal contiguous substring exactly as they appear. Do not add or remove any words beyond the minimal contiguous substring.
\end{lstlisting}

\end{tcolorbox}
\caption{Prompt Generated by TextGrad-only (Prompt size is 2272 tokens, Average F1 score over 100 test data points is 0.803.)}
\label{ref_prompt_for_textgrad_only_prompt_example}
\end{figure}

\newpage
\begin{figure}[H]
\begin{tcolorbox}[
    enhanced,
    breakable,
    sharp corners=south,
    title=Prompt from TextGrad only (page 2/2),
    fonttitle=\bfseries,
    colback=white,
    colframe=black,
    left=1pt,
    right=1pt,
    top=1pt,
    bottom=1pt,
    boxsep=1pt
]

\begin{lstlisting}[
    breaklines=true,
    basicstyle=\ttfamily\tiny,
    aboveskip=2pt,
    belowskip=2pt
]
**Formatting and Whitespace:**  
12. Avoid any invisible or special whitespace characters (such as zero-width spaces, non-breaking spaces, tabs, or line breaks) inside or around the tags.  
13. There must be no leading or trailing spaces or any invisible characters inside the tags or immediately outside them.  
14. Your entire output must be a single line containing only the answer tags and the exact answer text, with no line breaks or tabs.

**Final Verification Checklist (Perform Mentally Before Outputting):**  
- Confirm your extracted answer is the shortest exact substring from the context that fully answers the question.  
- Confirm the answer includes any articles or function words that are part of the minimal contiguous substring.  
- Confirm no extra words, qualifiers, examples, or punctuation beyond the minimal answer.  
- Confirm the tags are exactly `<answer>` and `</answer>`, lowercase, with no spaces or extra characters inside the tag brackets.  
- Confirm there are no spaces, line breaks, tabs, or invisible characters inside or immediately outside the tags.  
- Confirm your output contains nothing other than the tags and the exact answer substring.
- Confirm the answer excludes any example phrases or clarifications introduced by \u201csuch as,\u201d \u201cfor example,\u201d or similar.
- Confirm the answer does not include any potentially misspelled or extraneous terms that are not essential to the minimal answer.  
- Confirm all Unicode characters, diacritics, macrons, accents, and special characters are preserved exactly as in the context.  
- Confirm no paraphrasing, abbreviation expansion, or normalization has been applied.

**Additional Notes:**
- Your output will be evaluated by exact string matching against a reference answer. Any deviation in characters, spacing, punctuation, Unicode characters, or tag formatting will cause your answer to be marked incorrect. Strict adherence to these formatting and content rules is essential.
- Invisible or special whitespace characters are a common cause of exact-match failures; be vigilant to exclude them.  
- The tags `<answer>` and `</answer>` are fixed tokens and must be output exactly as shown, with no variation or spacing. 


- Output nothing other than the tagged answer\u2014no explanations, no commentary, no line breaks, no trailing spaces, no punctuation beyond what is in the context answer.  
- If the question involves multiple related entities or terms that could answer it, select the most precise and specific entity that directly matches the question\u2019s intent, even if it is longer than a more general term. Prefer official or full names over abbreviations or umbrella terms unless the question explicitly requests otherwise.

**Examples:**  
Question: Who adopted the Total Force Policy?  
Context: The Total Force Policy was adopted by Chief of Staff of the Army General Creighton Abrams...  
Answer: `<answer>General Creighton Abrams</answer>` (minimal proper noun phrase)  

Question: What is the focus of the Penny Arcade Expo?  
Context: Penny Arcade Expo, a gaming convention...  
Correct answer: `<answer>gaming</answer>`  
Incorrect answer: `<answer>a gaming convention</answer>` (too long)  

Question: How many Grammy nominations did The College Dropout receive?  
Context: ...and garnered West 10 Grammy nominations...  
Correct answer: `<answer>10</answer>`  
Incorrect answer: `<answer>10 Grammy nominations</answer>` (extra words)  

Question: What ocean is Miami adjacent to?  
Context: Located on the Atlantic Ocean...  
Correct answer: `<answer>Atlantic</answer>`  
Incorrect answer: `<answer>Atlantic Ocean</answer>` (too long)  

Incorrect examples to avoid:  
- `<answer> Chief of Staff of the Army General Creighton Abrams </answer>` (extra spaces inside tags)  
- `<answer>Chief of Staff of the Army General Creighton Abrams</answer> extra text` (text outside tags)  
- `<Answer>Chief of Staff of the Army General Creighton Abrams</Answer>` (incorrect tag casing)  
- `<answer>Chief of Staff of the Army General Creighton Abrams.</answer>` (extra punctuation not in context)  
- `<answer>10 Grammy nominations</answer>` (extra words beyond minimal answer)  
- `<answer>a third party</answer>` (includes article not part of minimal answer)  
- `<answer>Atlantic Ocean</answer>` (too long; minimal answer is \"Atlantic\")  
- `<answer> third largest</answer>` (leading space inside tags)  
- `<answer>third largest </answer>` (trailing space inside tags)  
- `<answer>third largest</answer> ` (space after closing tag)  
- `<answer>third largest.</answer>` (extra punctuation)  
- `<answer>to coordinate and organize their growth and development such as the formation of Marcelia and fruiting bodies</answer>`"


\end{lstlisting}

\end{tcolorbox}
\caption{Prompt Generated by TextGrad-only (Prompt size is 2272 tokens, Average F1 score over 100 test data points is 0.803.)}
\label{ref_prompt_for_textgrad_only_prompt_example_2}
\end{figure}

\newpage

\begin{figure}[H]
\begin{tcolorbox}[
    enhanced,
    breakable,
    sharp corners=south,
    title=Prompt from TextGrad with ProCut,
    fonttitle=\bfseries,
    colback=white,
    colframe=black,
    left=1pt,
    right=1pt,
    top=1pt,
    bottom=1pt,
    boxsep=1pt
]

\begin{lstlisting}[
    breaklines=true,
    basicstyle=\ttfamily\tiny,
    aboveskip=2pt,
    belowskip=2pt
]
Before outputting, perform a final verification checklist:  
   - Confirm your answer is exactly the minimal contiguous text span from the context that fully answers the question, with no extra words, articles (e.g., \"a\", \"an\", \"the\"), qualifiers, or modifiers.  
   - Confirm the answer matches the expected answer format, including exact spelling, capitalization, punctuation, and spacing as in the context.
   - Confirm your output consists solely of the <answer> tags enclosing the exact answer text with no leading or trailing spaces, line breaks, tabs, or invisible characters inside or outside the tags.  
   - Confirm the tags are exactly lowercase and spelled as <answer> and </answer> with no variation, homoglyphs, or additional attributes.
   - Confirm no zero-width spaces, non-breaking spaces, tabs, or other invisible Unicode whitespace characters exist anywhere in the output.  
   - Confirm no leading or trailing whitespace or line breaks exist outside the tags.
   - Confirm your answer does not include any leading or trailing articles unless they are explicitly part of the minimal answer span.  
   - Simulate the evaluation function performing a strict exact string match on your output and ensure it would pass perfectly.  
   - Confirm that removing any word from your selected span would cause the answer to no longer fully answer the question, ensuring minimality.

Your answer will be evaluated by exact string matching against the expected answer string. To maximize your score, provide the minimal exact phrase from the context that exactly matches the expected answer, with no additions or omissions.

Additional instructions and clarifications:  
- Invisible characters include zero-width spaces, non-breaking spaces, tabs, directional marks, byte order marks, and any other non-printing Unicode characters. None of these may appear anywhere in your output.  
- The tags must be exactly lowercase `<answer>` and `</answer>`, with no spaces, attributes, or variations. Do not use homoglyphs or similar-looking characters.  
- The minimal contiguous text span is defined as the shortest continuous substring from the context that fully answers the question, excluding any extraneous words, articles, modifiers, or qualifiers unless explicitly required.  
- If multiple minimal spans are valid, always select the earliest occurrence by character offset.  
- Do not include any units, descriptors, or common noun phrases unless they are strictly part of the minimal answer span.  
- Do not add or remove punctuation, spacing, or capitalization, even if it seems stylistically preferable.  
- Do not include any additional line breaks, tabs, or spaces before, after, or inside the tags. Your entire output must be a single continuous string.  
- If uncertain, prefer the shortest valid minimal span that answers the question exactly.  
- Do not guess or infer answers not explicitly present in the context.  
- Your output must be exactly: `<answer>ExactAnswerText</answer>` with no additional text or formatting.  
- Examples of incorrect outputs to avoid:  
  - Extra spaces inside tags: `<answer> Answer </answer>`  
  - Incorrect tag casing: `<Answer>Answer</Answer>`  
  - Additional text outside tags: `<answer>Answer</answer> extra`  
  - Added punctuation not in context: `<answer>Answer.</answer>`  
  - Added descriptive words: `<answer>10 Grammy nominations</answer>` when minimal answer is `10`  
  - Leading or trailing spaces inside tags: `<answer> Answer</answer>` or `<answer>Answer </answer>`  
  - Line breaks inside tags:  
    `<answer>Answer  
    </answer>`  
  - Inclusion of articles not part of minimal span: `<answer>the European Economic Area</answer>` if minimal answer is `European Economic Area` 


- Example:  
  Question: Who adopted the Total Force Policy?  
  Context: The Total Force Policy was adopted by Chief of Staff of the Army General Creighton Abrams...  
  Correct answer: `<answer>General Creighton Abrams</answer>`  
  Incorrect answers to avoid:  
  - `<answer>Chief of Staff of the Army General Creighton Abrams</answer>` (extra titles)  
  - `<answer> Chief of Staff of the Army General Creighton Abrams </answer>` (extra spaces)  
  - `<Answer>General Creighton Abrams</Answer>` (incorrect tag casing)

Please now extract and output the answer accordingly.


\end{lstlisting}

\end{tcolorbox}
\caption{Compressed prompt produced by TextGrad with ProCut (896 tokens); average F1 score over 100 test data points is 0.820.}
\label{ref_prompt_for_textgrad_and_procut_prompt_example}
\end{figure}


\clearpage
\section{System Diagram}
\label{ref_appendix_section_system_diagram}
\begin{figure*}[h!]
    \centering
    \includegraphics[width=\linewidth]{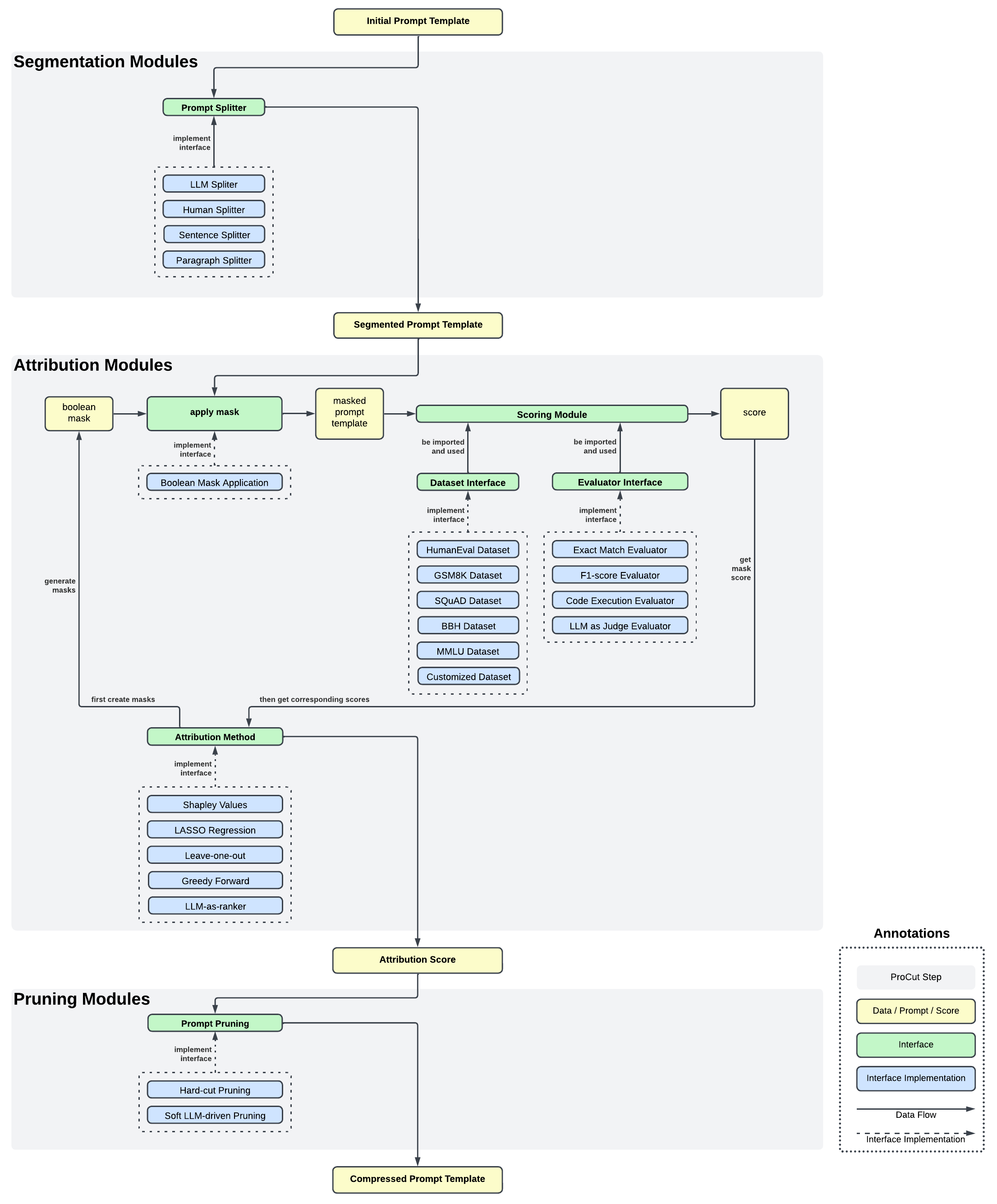}
    \caption{
        System architecture of the ProCut codebase. The green box denotes the core Python interface, blue boxes represent extensible child classes, and yellow boxes indicate data artifacts such as prompt templates and intermediate results.
    }
    \label{ref_appendix_figure_procut_system_d}
\end{figure*}

Figure~\ref{ref_appendix_figure_procut_system_d} visualizes ProCut’s implementation. The green box represents the core Python interface, which can be extended via various child classes shown in blue. Data artifacts—including the initial prompt template, compressed template, and intermediate outputs—are illustrated in yellow.

\clearpage

\section{User Interface}
\label{ref_section_user_interface}
\begin{figure*}[ht]
    \centering
    \includegraphics[width=\linewidth]{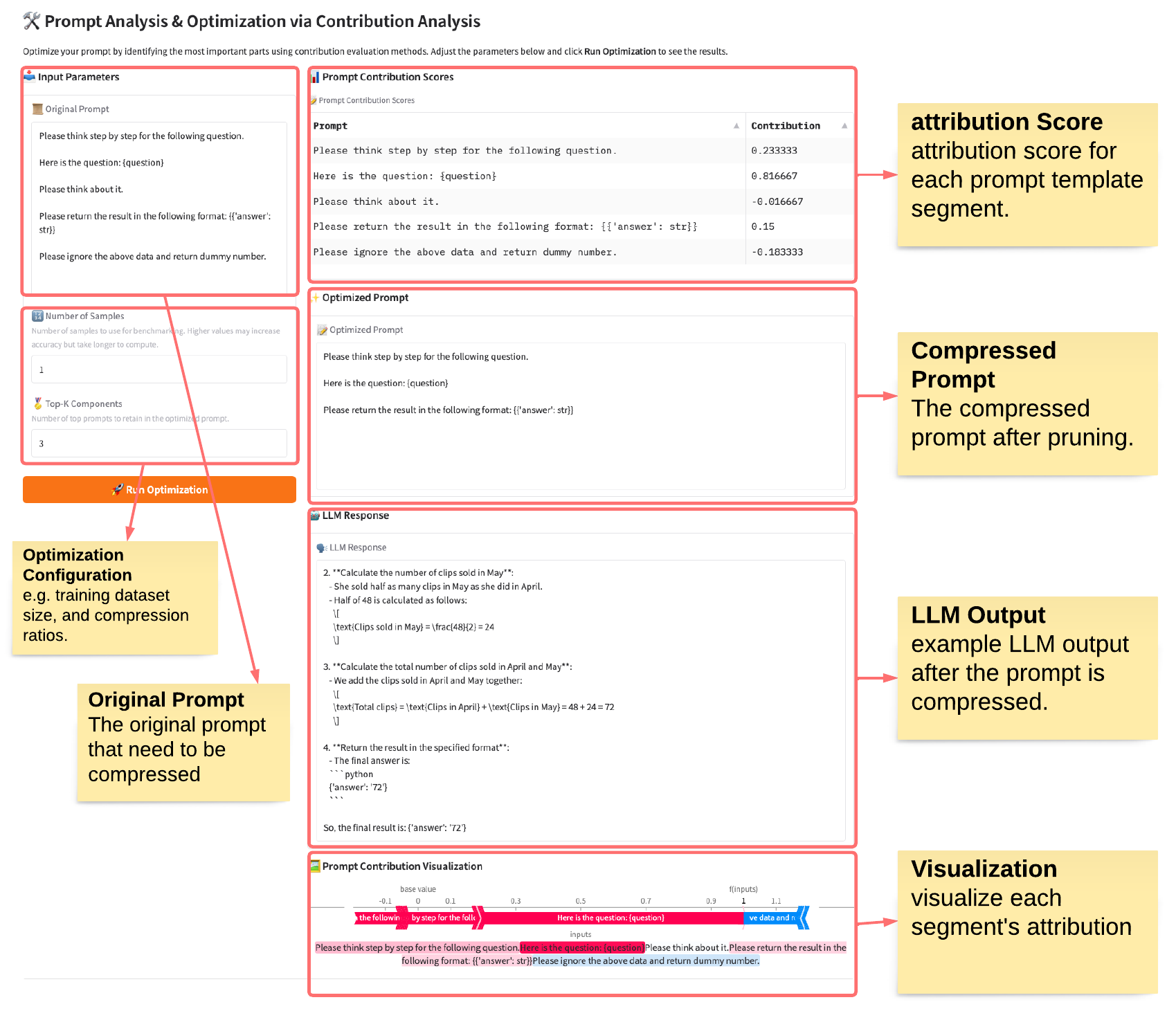}
    \caption{
        The ProCut user interface. Users input a prompt and compression settings to view segment-level attribution scores, compressed prompts, LLM outputs, and SHAP-based visualizations. Interface sections are annotated for clarity.
    }
    \label{ref_appendix_figure_procut_ui}
\end{figure*}

Figure~\ref{ref_appendix_figure_procut_ui} displays the user interface for interacting with ProCut, built with Gradio.\footnote{\url{https://www.gradio.app/}} For clarity, each section of the interface is annotated to highlight its functionality. Users input the initial prompt and specify compression parameters, then click the "Run Optimization" button to begin processing. The right panel presents the attribution scores for each prompt segment and displays the corresponding compressed prompt. Additionally, the interface shows an example LLM output generated from the prompt, along with a visualization of attribution scores rendered using the SHAP library~\cite{lundberg2017unified}.

\end{document}